\pdfoutput=1
\documentclass[11pt]{article}
\usepackage[utf8]{inputenc}
\usepackage[margin=1in]{geometry}
\usepackage{amsmath,amssymb}
\usepackage{booktabs}
\usepackage[section]{placeins}
\usepackage{graphicx}
\usepackage{xcolor}
\usepackage{tikz}
\usetikzlibrary{arrows.meta,positioning,fit,backgrounds}
\usepackage[numbers,sort&compress]{natbib}
\usepackage[pdftitle={Focus Only on What's Important: Training-Free Focus Maps for Frozen Vision Transformers in Long-Video Object Segmentation},pdfauthor={Harsha Karipineni, Rajesh Reddy Chintha},colorlinks=true,linkcolor=blue!50!black,citecolor=blue!50!black,urlcolor=blue!50!black]{hyperref}
\usepackage{microtype}
\usepackage{enumitem}
\setlist{itemsep=2pt,topsep=3pt}

\newcommand{\jf}{$\mathcal{J}\&\mathcal{F}$}
\newcommand{\ci}[2]{\mbox{\small$[#1,\,#2]$}}

\title{Focus Only on What's Important: Training-Free Focus Maps for Frozen Vision Transformers in Long-Video Object Segmentation}
\author{Harsha Karipineni\\ Independent Researcher\\ \texttt{boletoms@gmail.com}
\and Rajesh Reddy Chintha\\ Independent Researcher\\ \texttt{chintharr@gmail.com}}
\date{}

\begin{document}
\maketitle

\begin{abstract}
Vision transformers are strong but costly image encoders for video object segmentation. Training-free token merging cuts compute uniformly across the frame, including on tracked objects. We give a frozen DINOv3 ViT inside a trained Cutie head a training-free focus map: full resolution where the segmenter expects its objects and coarser merging elsewhere. Protecting only the previous frame's mask works on short clips but fails on long videos, where fast motion pushes targets outside the margin and into background merging, locking in loss. Our policy instead takes the union of velocity-shifted and block-matched masks, runs a camera-compensated surprise check, delegates uncertain regions to the compared merger, and falls back to it when all objects are lost. On long videos (LVOS V2 valid, pre-specified protocol, three ViT-S seeds), it beats ALGM~0.98 by $+2.76$ \jf{} (95\% object CI $[+1.22,+4.44]$; video-cluster $[+0.99,+4.52]$) at 16\% fewer counted encoder FLOPs ($0.385$ vs.\ $0.459$), sitting $1.0$ \jf{} point below the dense-encoder baseline; at ViT-B the gain over ALGM~0.95 is $+3.89$ $[+2.03,+5.87]$. On DAVIS 2017 val it stays within $-0.19$ \jf{} of the dense encoder at ViT-S ($-0.12$ at ViT-B). Controls qualify the result: protecting only the previous mask under ALGM matches or exceeds accuracy at higher FLOPs ($0.486$), so expectation and surprise mainly reduce counted encoder compute. Practically, lower counted encoder compute did not reduce startup-inclusive mean Cutie step time on a 12-video LVOS sample (3\%--8\% lower in steady state; an equally engineered ALGM is faster at ViT-S and ViT-B). Section~\ref{sec:full} separately examines a Cutie permanent-memory schedule and its latency and memory trade-offs. Code: \url{https://github.com/boletoms/focused-cv}
\end{abstract}

\section{Introduction}

Frozen self-supervised vision transformers such as DINOv3~\cite{simeoni2025dinov3} give strong dense features, which makes them attractive encoders for video object segmentation (VOS). Their cost grows with the number of patch tokens, and VOS processes every frame at high resolution. Training-free token reduction~\cite{bolya2023tome,norouzi2024algm,li2023ailurus,tran2024pitome,liang2022expedite} lowers that cost by merging similar tokens, with one rule for the whole image.

VOS offers something these methods do not use: the segmenter's prediction for the previous frame says where the objects are. A natural design keeps the predicted objects, plus a margin, at full resolution and merges the rest hard. We call the per-patch map that decides how much compute each patch gets a \emph{focus map}, and its full-resolution patches the \emph{focus areas}. On short DAVIS clips the closed-loop focus map works: it comes within 1 \jf{} of dense at 0.29 of dense FLOPs, half of what ALGM 0.98 uses (Section~\ref{sec:short}). On long videos it fails. Our diagnostics (Section~\ref{sec:why}) show why: when a target moves more than two patches between frames, part of it falls outside the margin and is merged with the background; the target is lost, and since the map follows the prediction, it stays lost. On such frames the closed-loop map loses targets nine times as often as dense; away from fast motion it loses them no more often than dense.

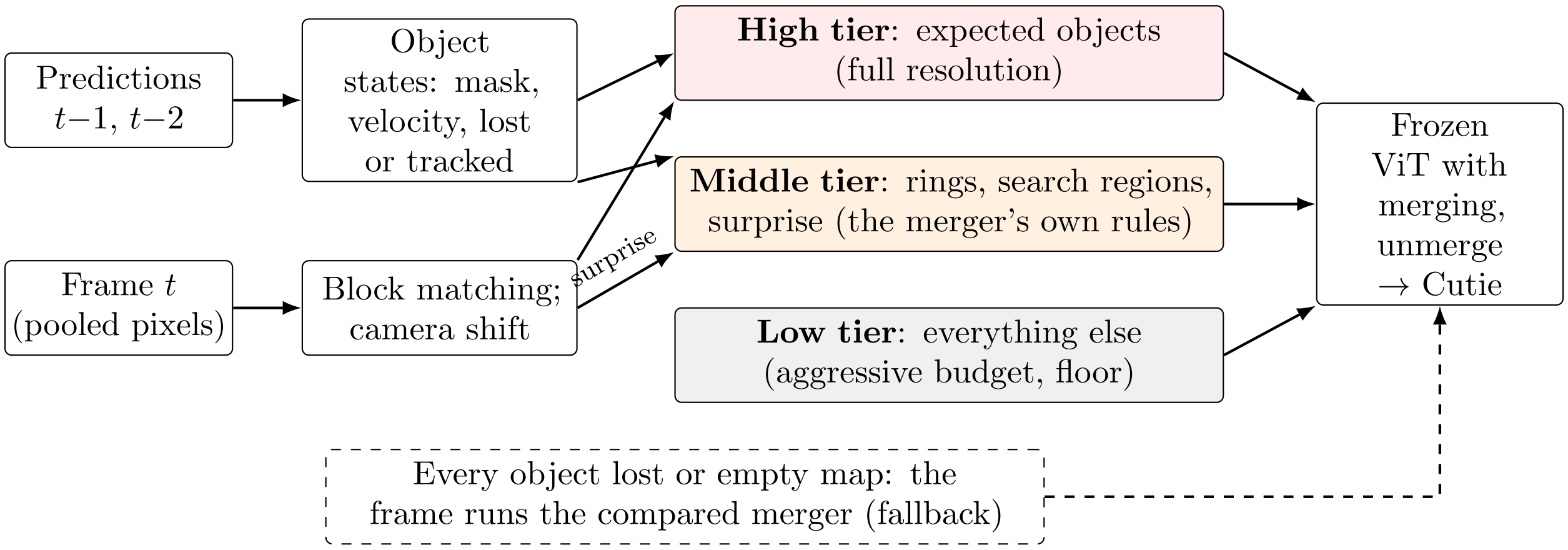
\begin{figure}[htbp]
\centering
\resizebox{\linewidth}{!}{%
\begin{tikzpicture}[font=\small,
  box/.style={draw, rounded corners=2pt, align=center, minimum height=10mm, inner sep=3pt, fill=white},
  arr/.style={-{Latex[length=2mm]}, thick}]
\node[box, text width=22mm] (prev) at (0, 0) {Predictions\\$t{-}1$, $t{-}2$};
\node[box, text width=22mm] (frame) at (0, -2.2) {Frame $t$\\(pooled pixels)};
\node[box, text width=27mm] (state) at (3.4, 0) {Object states: mask,\\velocity, lost or tracked};
\node[box, text width=27mm] (bm) at (3.4, -2.2) {Block matching;\\camera shift};
\node[box, text width=56mm, fill=red!8] (high) at (8.8, 0.5) {\textbf{High tier}: expected objects\\(full resolution)};
\node[box, text width=56mm, fill=orange!12] (mid) at (8.8, -1.1) {\textbf{Middle tier}: rings, search regions,\\surprise (the merger's own rules)};
\node[box, text width=56mm, fill=gray!12] (low) at (8.8, -2.7) {\textbf{Low tier}: everything else\\(aggressive budget, floor)};
\node[box, text width=24mm] (vit) at (14.0, -1.1) {Frozen ViT with\\merging, unmerge\\$\rightarrow$ Cutie};
\node[box, text width=74mm, dashed] (fb) at (6.0, -4.2) {Every object lost or empty map: the frame runs the compared merger (fallback)};
\draw[arr] (prev) -- (state);
\draw[arr] (frame) -- (bm);
\draw[arr] (state.east) -- (high.west);
\draw[arr] (state.south east) -- (mid.north west);
\draw[arr] (bm.north east) -- (high.south west);
\draw[arr] (bm.east) -- node[above, sloped, font=\scriptsize, pos=0.5]{surprise} (mid.south west);
\draw[arr] (high.east) -- (vit.north west);
\draw[arr] (mid.east) -- (vit.west);
\draw[arr] (low.east) -- (vit.south west);
\draw[arr, dashed] (fb.east) -| (vit.south);
\end{tikzpicture}}
\caption{The focus map. Before the encoder runs on frame $t$, for each tracked object we take the union of its velocity-shifted and block-matched masks on frame $t$ (high tier). A surprise check compares frame $t$ with frame $t{-}1$ shifted by the camera motion and promotes patches that changed without explanation. Rings sized by each object's speed and search regions around lost objects form the middle tier, merged by the compared merger's own rules; the rest is merged aggressively. Every decision is made before the first block, on the patch grid.}
\label{fig:method}
\end{figure}

We propose to stop trusting the last mask: expect where each object will be, then check the frame against that expectation. The method has four parts, all training-free (Figure~\ref{fig:method}):
\begin{enumerate}
\item \textbf{Expectation.} For each tracked object, take the union of its velocity-shifted and block-matched masks (found on the current frame's pooled pixels), grown by one patch (two at ViT-B). This region runs at full resolution.
\item \textbf{Surprise check.} A per-patch mismatch between the current frame and the previous frame shifted by the estimated camera motion, with a variance floor and a small spatial tolerance. Patches that fire are promoted, never demoted.
\item \textbf{Three tiers.} The expectation (high tier) is never merged. Speed-scaled rings, search regions around lost objects and surprise patches (middle tier) are merged by the compared merger's own rules. Everything else (low tier) is merged by an aggressive budget with a floor of context tokens.
\item \textbf{Fallback.} When every object is lost, the frame runs the compared merger unchanged.
\end{enumerate}

Our contributions are:
\begin{itemize}
\item A diagnosis of why closed-loop focus maps fail on long videos: fast motion causes compression-induced losses that the loop locks in.
\item A training-free focus policy that runs at inference time in a trained Cutie head receiving stride-16 features from frozen DINOv3 (tested with one segmenter, Cutie, and one backbone family, DINOv3), conditioning compute on an expectation of object and camera motion, a surprise check, and a fallback to the compared merger.
\item A pre-specified evaluation: the rule for every adoption decision was written in dated plan files before the run that tested it, and the long-video test pools three heads (one base head fine-tuned with three seeds), at ViT-S and again at ViT-B. We report every run, including failures.
\item Ablations showing that full resolution in the expected region is necessary, and an exact attribution of where the long-video gain comes from.
\item A training-free permanent-memory schedule for Cutie (a finding about the segmenter, not the focus map) and, with it, operating points at which the encoder runs at about a third of dense FLOPs at or above dense accuracy on held-out long videos (Section~\ref{sec:full}).
\item A graph-compatible implementation that takes the same decisions as the reference implementation and turns the FLOPs saving into wall-clock speed-ups over dense, with the same engineering applied to the strongest baseline.
\end{itemize}

\section{Related work}

\paragraph{Training-free token reduction for dense prediction.} ToMe~\cite{bolya2023tome} merges the most similar token pairs by bipartite matching in every block. ALGM~\cite{norouzi2024algm} merges locally in $2\times2$ windows early and globally later with a similarity threshold, designed for segmentation with plain ViTs. AiluRus~\cite{li2023ailurus} adapts resolution to content, PiToMe~\cite{tran2024pitome} preserves the spectrum of the token graph, and Expedite~\cite{liang2022expedite} clusters tokens and reconstructs a dense output. G2TM~\cite{g2tm} coarsens a graph of spatially adjacent tokens and SCMerge~\cite{scmerge} penalises pair merges that break spatial consistency, both for semantic segmentation; DTEM~\cite{lee2024dtem}, unlike the others, learns its merging embeddings. Each applies one rule to the whole image. We port ToMe, ALGM, AiluRus, PiToMe and Expedite from their authors' code and compare them inside the same VOS model at operating points fixed in advance (Section~\ref{sec:setup}). Content-aware token sharing~\cite{lu2023cts} lets image regions share tokens by content and dynamic token pruning~\cite{tang2023dtop} drops easy tokens in plain-ViT segmenters, both image-level. MPM~\cite{rave2026mpm} pairs mutual nearest neighbours and CubistMerge~\cite{gong2025cubist} keeps the spatial layout so that merging works with window attention and rotary embeddings; both are image-level and use no information from earlier frames.

\paragraph{Prompt- and region-aware reduction.} Segment This Thing~\cite{schmidt2025stt} tokenizes finely around a point prompt and coarsely elsewhere, the closest image-level analogue of a focus map, without a previous frame; StructSAM~\cite{structsam} protects boundary and prompt regions when merging in SAM; PrATo~\cite{prato} prunes by a prompt-derived spatial prior; VLTP~\cite{vltp} prunes by language guidance. In video, for SAM 2~\cite{ravi2025sam2}, Fast SAM2~\cite{fastsam2} prunes encoder outputs by text before propagation; Efficient-SAM2~\cite{efficientsam2} keeps full encoder computation in the windows covered by the previous frame's dilated predicted masks, adds salient windows when tracking confidence is low, and routes the rest through a light branch trained by feature reconstruction; the Recurrent Dynamic Submodel~\cite{tang2026rds} chooses SAM 2's blocks per frame with a learned router that reads the previous mask and the current features. Motion-aware ROI prediction~\cite{fu2021motionroi} forecasts a target's crop from its trajectory. Video Patch Pruning~\cite{vpp} learns to select patches early from the previous frame's features for video instance segmentation, and MaskVD~\cite{maskvd} masks regions for video detection without training, refreshing full frames periodically. RS$^3$-Prune~\cite{rs3prune} prunes the readout queries and memory writes of memory-bank VOS without training. Our closed-loop focus map belongs to the family that uses the previous prediction as the prompt. The method we propose differs in where the full-resolution region comes from (an expectation of object and camera motion, corrected on the current frame), in its surprise check, in giving an uncertain middle tier to the compared merger's own rules, in falling back to that merger when tracking fails, and in its analysis of long videos, where we find that the plain closed-loop map fails.

\paragraph{Temporal reuse.} Eventful Transformers~\cite{dutson2023eventful} and related methods reuse computation for unchanged tokens across frames, and temporal cluster assignment~\cite{tca} reuses token clusters. Run-length tokenization~\cite{choudhury2024rlt} drops patches repeated over time and names camera motion as its failure mode, and vid-TLDR~\cite{choi2024vidtldr} merges background tokens of video transformers by attention-based saliency. A reuse option we built in the same codebase, in the style of Eventful Transformers (our implementation, not the authors' code; skipped patches keep their cached output and serve as keys and values), saved at most 2\% of FLOPs within one \jf{} point of the budget alone on DAVIS 2017 val (30 videos, closed-loop maps), and 10 to 15\% at a cost of 1.2 to 1.4 points: after the budget, low-interest areas are merged to a floor, and handheld camera motion changes almost every patch. It was not run on long videos or with the expectation focus map. We reduce tokens within each frame; the focus map uses the previous frames only to decide where. Long-video segmenters address memory rather than encoder cost: XMem~\cite{cheng2022xmem} and Cutie~\cite{cheng2024cutie} keep long-term memories, and SAM2Long~\cite{ding2025sam2long} keeps several memory pathways without training. Pollard and Wray~\cite{pollard2025video} study training-free token merging in video transformers, which merge tokens across space and time with one rule for the whole clip, and report about 2.5 times the speed at a small accuracy cost on action recognition.

\paragraph{Memory-based VOS.} Cutie~\cite{cheng2024cutie} reads object-level memory to segment each frame; LVOS~\cite{hong2023lvos} measures long-term VOS, where losses and re-appearances dominate. We keep Cutie unchanged except its pixel encoder, whose stride-16 features come from the frozen ViT under each method; its memory readout and decoder are identical for every method and excluded from the compute we report. For SAM 2, SAM2Dual~\cite{sam2dual} adds a training-free long-term memory filled by interval sampling to a short-term one, a memory schedule in the spirit of Appendix~\ref{app:perm} but for a different segmenter and without an encoder-compute saving; and the codec-guided compressor of Sheng et al.~\cite{sheng2026codec}, for video language models, uses compressed-video motion vectors as a prior for merging tokens across frames, which is temporal and training-free but not tied to a segmenter's predicted objects. SAMURAI~\cite{yang2024samurai} adds a training-free Kalman motion model to SAM 2's memory selection for tracking, and EdgeTAM~\cite{zhou2025edgetam} and EfficientTAM~\cite{xiong2024efficienttam} replace SAM 2's image encoder by lighter trained ones and also make its memory attention cheaper; none of them changes the encoder's compute per frame from the segmenter's predictions.

\section{Method}

\subsection{Setting}
The encoder is a frozen DINOv3 ViT (ViT-S/16; ViT-B/16 and ViT-L/16 as checks) with 4 register tokens and RoPE positions. A method may merge patch tokens between attention and MLP in any block. Each token carries its size (how many original patches it holds), a size-weighted position used for RoPE, and its tier. After the last block every original patch takes the output of the token that holds it (unmerge), so Cutie's decoder receives a full stride-16 feature map. The ViT is never trained; Cutie's remaining modules are trained once per head with the ViT dense (Section~\ref{sec:setup}).

\subsection{Why a closed-loop focus map fails on long videos}\label{sec:why}
The closed-loop focus map protects every patch the previous prediction touches, grown by one patch (two at ViT-B), and merges the rest with a budget: in every block bipartite matching merges the remaining low-tier tokens among themselves at a nominal rate of 0.7, which pairing caps at one half, and never below 64 of them remain. It comes within 1 \jf{} of dense on DAVIS at 0.29 of dense FLOPs (Table~\ref{tab:davis_s}) but falls below ALGM on long videos (paired differences from $-5.38$ to $-0.10$ \jf{} across the heads we tested on LVOS V1 val).

Logged runs on the development split (LVOS V2 train) locate the cause. The share of a tracked target outside the protected patches grows with its displacement since the previous frame: 0.6\% under one patch, 12.1\% at 2 to 4 patches, 31.5\% beyond 4 patches. LVOS runs at 6 frames per second, so this lag is geometric and the same for every map source, including the previous ground truth. On frames where a target moves more than two patches (about 8\% of tracked frames), the closed-loop map loses it on 2.81\% of frames, against 0.30\% for dense and 0.36\% for ALGM (each rate counts the frames on which that method was still tracking the target, so the three come from different sets of frames); on all other frames the three are equal (0.01 to 0.03\%). Ninety-six percent of its loss onsets fall on fast frames. Losses are not always permanent (92\% of the closed-loop map's visible-loss onsets recovered, 741 of 809, on the development split), but the deficit against ALGM sits in the frames where the closed-loop map is lost and ALGM tracks. The losses are compression-induced (the lagging part of the target is merged with background), and the loop locks them in. Remedies tried on the closed-loop map, a dense I-frame every 5 frames and dense I-frames triggered by evidence of the target outside the map, helped on development videos but not on LVOS V1 val; this led to the design below, which replaces full resets by an expectation, a surprise check and the fallback.

\subsection{Expectation (high tier)}\label{sec:expect}
For each object we maintain a run-time state: its last mask, centroid, and velocity estimated from predictions at $t{-}1$ and $t{-}2$, together with a separate history of up to 10 tracked-mask areas. In the runtime controller, an object counts as lost when its predicted mask area falls below a tenth of the median of those last 10 tracked areas, or below 16 pixels (distinct from the ground-truth IoU $<0.1$ loss definition used in the outcome analysis of Section~\ref{sec:attr}). For each tracked object, we take the union of the velocity-shifted and block-matched masks---(a) its last mask shifted by its centroid velocity and (b) its last mask shifted by the displacement that best matches its template between the previous and the current frame on $4\times4$-pooled pixels (search within 8 patches)---and grow that union by one patch (two at ViT-B). Object masks are taken at patch level (a patch belongs to the mask at 1\% coverage; the template uses the patches covered at least half, or the mask if there are none). The high tier is the union over objects, plus the previous prediction and its velocity shift; before the expected regions are added, the base map from the previous prediction is rounded up to the next multiple of 64 patches with its nearest unprotected patches (at most 63 extra, as for the closed-loop map). The high tier is never merged, although its tokens still attend to merged ones. Offline on 420 development videos, this expectation covers 93.3\% of fast-moving targets against 77.8\% for the one-patch margin, with 31\% more area on those frames (10\% over all frames); it was chosen by a rule fixed before that measurement.

\subsection{Surprise check}
A surprise is any deviation from the expected frame. For each patch $p$ outside the expectation we compare the current frame's pooled pixels at $p$ with the previous frame's at $p-g$, where $g$ is the integer camera shift (within 4 patches) that best aligns the two frames. The mismatch is normalised per patch with a variance floor $\sigma=0.05$ and takes the best of nine offsets of up to 4 pixels; a patch fires at $e\geq0.06$, the highest threshold that reached 95\% recall on both offline event sets (targets entering outside the map, and re-appearances). Content entering at the image border always fires. A fired patch inside an object's ring or search region joins the high tier; any other fired patch joins the middle tier. If more than half of the patches outside the base map fire (a frame-wide change such as a cut or fast camera motion), all of them join the middle tier while the expectation stays whole. Surprise only promotes: false positives cost compute, false negatives cost accuracy, and we prefer the former.

\subsection{Three tiers}
\begin{itemize}
\item \textbf{High:} the expectation and promoted surprise patches; full resolution.
\item \textbf{Middle:} a ring of $1+\max(1,\lceil s\rceil)$ patches around each expectation ($s$: speed in patches per frame); for each lost object, its last tracked mask moved by $n$ times its last velocity and grown by $2+n$ patches, $n$ frames after the loss; surprise patches not in the high tier. Merged by the compared merger's own rules: for ALGM, its layer-1 $2\times2$ windows (only windows without a high-tier patch) and its threshold matching between non-high tokens in layer 5 (layer 7 in the 24-block ViT-L).
\item \textbf{Low:} everything else, merged by ALGM's rules where they apply and then by the budget (nominal rate 0.7 per block, capped at one half by pairing; floor 64 tokens at ViT-S, 128 at ViT-B and ViT-L). A merge of middle and low tokens stays middle, so the budget never reaches the middle tier.
\end{itemize}
All tier decisions are made before the first merge, while every patch is still a token. The focus areas are the high- and middle-tier patches.

\subsection{Fallback}
When every object that has appeared is lost, or the map is empty, the frame runs the compared merger exactly (plain ALGM 0.98 at ViT-S); the expectation observes passively so its state stays current. On the held-out protocol the fallback runs on 29.6\% of frames (three heads); Section~\ref{sec:attr} shows the gain is not produced by these frames.

\subsection{Compute accounting}
We count whole-encoder FLOPs per frame from what the forward pass did: projections on the tokens entering attention, attention per computed query--key pair, MLP on the tokens left after merging, the patch embedding, and every operation a method adds (matching, clustering, and our feature products for the camera search, block matching and the surprise mismatch). Norms, softmax and RoPE are omitted for every method alike. We report the ratio to dense, averaged per frame (frame-weighted across videos); dense frames inside a method's run count 1.0. Cutie's memory and decoder are identical across methods and excluded. The count follows the tokens a method keeps; the timed implementation's padded buffers and redone blocks are overhead of that implementation and appear only in wall clock. Wall clock is measured separately (Section~\ref{sec:time}).

\subsection{Timed implementation}
A direct implementation is bound by the host: per-frame Python work, host reads of token counts and many small kernels outweigh the work it removes. With the method unchanged, the timed implementation keeps every step before the encoder on the GPU (two host reads per frame), sizes each block's token buffer from the previous frame's counts (a frame that exceeds a buffer is redone from that block on, so outputs never depend on the guess), captures every step in CUDA graphs, compiles each block segment, uses variable-length flash attention~\cite{dao2024flashattention2}, and overlaps block 0's attention with the tier planning. It equals its own eager form bit for bit, rounds every bf16 intermediate as the reference path does, and takes the reference path's decisions: replaying the ground-truth masks of all 140 long videos through both, the fallback decision agrees on every frame and the tiers on all but one of 61,880 frames (an fp32 centroid at exactly half a pixel; Appendix~\ref{app:impl}). Every number of the method in Tables~\ref{tab:long}--\ref{tab:time} comes from this implementation unless marked; baselines run their ported code.

\section{Experimental setup}\label{sec:setup}

\paragraph{Model.} Cutie (small) with its pixel encoder's stride-16 features replaced by the frozen DINOv3 ViT under each method; the stride-8 and stride-4 skips come from Cutie's own ResNet-18 stem. Heads (Cutie's modules other than the ViT) are trained with the ViT frozen and dense on DAVIS 2017 train and MOSE train~\cite{ding2023mose} with Cutie's training script; we do not train on YouTube-VOS or static images ourselves. Each head starts from the official Cutie small weights (\texttt{cutie-small-mega}; all non-ViT tensors except the two new stride-16 projections), which their authors trained with static-image pre-training and on DAVIS, YouTube-VOS, MOSE, BURST and OVIS, so our heads inherit that training; trains the projections for 5,000 iterations and then everything for 40,000 (batch 2). The long-video heads are fine-tuned from one such base head for 10,000 iterations with three Cutie seeds, so they share the base head's 45,000 iterations: R31c, h2 and h3 at ViT-S, and R41 b1 to b3 (from head R25) at ViT-B. ViT-L has one head.

\paragraph{Data.} DAVIS 2017 val~\cite{pont2017davis} (30 videos, 61 objects, 480p) for short clips. For the held-out long-video test, LVOS V2 valid~\cite{hong2025lvos2}: 140 videos with 239 objects, made of LVOS V1 val~\cite{hong2023lvos} (50 videos, 65 objects; its frames and annotations are identical in V2) and 90 videos new in V2 (174 objects). The method's components and settings were chosen on LVOS V2 train; the 50 V1 videos had, however, been used earlier to evaluate the closed-loop design and its variants (whose failure there motivated this work), so only the 90 new videos were untouched before this test, and we report them separately. Throughout, ``held-out'' means not used to choose the method's components or thresholds in the first place; LVOS V2 valid was nevertheless evaluated in many pre-specified passes (Appendices~\ref{app:impl} and~\ref{app:official} list them), and later choices made after the main result (the surprise threshold 0.5, the schedule operating points) were proposed after earlier held-out results had been seen. LVOS V2 train (420 videos) for development: diagnostics, gates and offline measurements. Cutie's LVOS protocol: shortest side 480, long-term memory, each object's mask given at its first appearance.

\paragraph{Metrics.} \jf{} per object, under two scorers. Ours scores every annotated frame after an object's first appearance at $1280\times720$, with a boundary tolerance of 0.8\% of the diagonal; all decisions in this paper were made with it. The official toolkits~\cite{hong2023lvos,pont2017davis}, run unchanged on the saved masks, score each object within its annotated range without its first and last frame (LVOS) or every frame but the first and last (DAVIS). We report both.

\paragraph{Baselines.} ToMe, ALGM, AiluRus, PiToMe and Expedite, ported from the authors' code at recorded commits and tested against the originals. Map-aware ``protected'' variants of ALGM keep the closed-loop focus map at full resolution and apply ALGM elsewhere. Operating points follow rules fixed in advance (for example ALGM's threshold closest to a target FLOPs on DAVIS: 0.98 at ViT-S, 0.95 at ViT-B and ViT-L).

\paragraph{Protocol and statistics.} Paired differences per object with 95\% bootstrap intervals (1,000 resamples of objects). Objects of one video share frames and tracker state, so Appendix~\ref{app:official} also reports intervals that resample whole videos (10,000 resamples) and the mean that weights videos equally, as sensitivity checks; the decision rule fixed in advance uses the object bootstrap. The long-video test was fixed before any run: per object, the difference against ALGM averaged over the three heads, then bootstrapped; superiority if the lower bound is above 0 at lower FLOPs, non-inferiority if above $-3.0$. This primary comparison is not adjusted; the secondary comparisons (per head, per split, and the method against each short-clip baseline) are reported with Holm-adjusted~\cite{holm1979} two-sided p-values per family from a null-centred bootstrap over whole videos (10,000 resamples; Appendix~\ref{app:official}). Every adoption rule (expectation variant, surprise threshold, gates, validation) was written in the project's dated plan files before the run that tested it (pre-specified, not registered externally). Evaluations use deterministic GPU kernels where PyTorch provides them; the timed implementation's compiled kernels are not bitwise reproducible across runs (Appendix~\ref{app:official}). Attention runs in bf16 or fp32, never fp16.

\section{Results}

\subsection{Long videos, held out}\label{sec:long}

\begin{table}[htbp]
\centering
\caption{Held-out long videos (LVOS V2 valid: 140 videos, 239 objects), ViT-S, three heads. FLOPs: encoder FLOPs as a share of dense. Differences: ours minus the row, per object, averaged over heads, 95\% bootstrap intervals. Official: the object mean of the LVOS toolkit's scores on a rerun that saved the masks; the method's rows differ between the two runs (our scorer gives 68.43 on the rerun, dense and ALGM are unchanged; Appendix~\ref{app:official}).}
\label{tab:long}
\begin{tabular}{lcccc}
\toprule
& & \multicolumn{2}{c}{\jf{}} & \\
\cmidrule(lr){3-4}
Method & FLOPs & our scorer & official & Ours $-$ method (our scorer) \\
\midrule
Dense & 1.000 & 69.47 & 69.19 & $-1.00$ \ci{-2.27}{+0.31} \\
ALGM 0.98 & 0.459 & 65.72 & 65.40 & $\mathbf{+2.76}$ \ci{+1.22}{+4.44} \\
\textbf{Ours} & \textbf{0.385} & \textbf{68.48} & \textbf{68.16} & \\
\bottomrule
\end{tabular}
\end{table}

\begin{figure}[htbp]
\centering
\includegraphics[width=\linewidth]{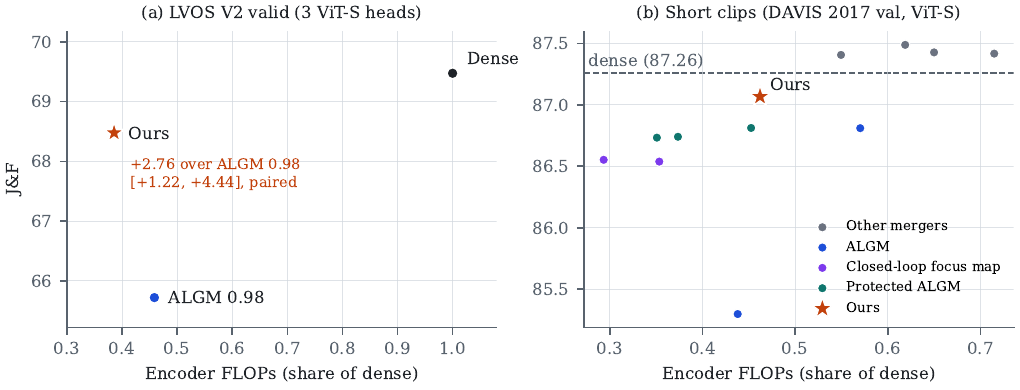}
\caption{\jf{} against encoder FLOPs. (a) LVOS V2 valid, mean over three ViT-S heads; the paired difference against ALGM 0.98 with its 95\% interval is given as text. (b) Short clips (the Table~\ref{tab:davis_s} baseline grid at ViT-S); the line marks dense.}
\label{fig:flops}
\end{figure}

Table~\ref{tab:long} gives the pre-specified test (Figure~\ref{fig:flops}a): the method is superior to ALGM 0.98 at lower FLOPs. Per head the difference is $+3.16$ \ci{+0.97}{+5.42}, $+2.68$ \ci{+0.27}{+5.01} and $+2.42$ \ci{-0.49}{+5.35}. Per split it is $+3.62$ \ci{-0.30}{+7.26} on LVOS V1 val (65 objects) and $+2.43$ \ci{+0.69}{+4.13} on the 90 new videos; on the new videos the method is below dense ($-1.95$ \ci{-3.59}{-0.52}). Resampling whole videos instead of objects, the difference against ALGM 0.98 is $+2.76$ \ci{+0.99}{+4.52}, and weighting every video equally gives $+2.81$ \ci{+0.73}{+5.00}; on the 90 new videos these are $+2.43$ \ci{+0.54}{+4.25} and $+1.65$ \ci{-0.33}{+3.69}. The estimate is smaller under equal video weighting, and its interval includes zero (Appendix~\ref{app:official}). Under the official LVOS evaluation the difference against ALGM is $+2.75$ \ci{+1.23}{+4.55}: $+2.55$ \ci{+0.87}{+4.40} on the 202 objects of seen categories and $+3.85$ \ci{-0.77}{+8.67} on the 37 of unseen ones. On the development split (420 LVOS V2 train videos, one head) the method is $+1.33$ \ci{+0.53}{+2.16} over ALGM 0.98 at 0.377 against 0.443 FLOPs, and $+3.79$ \ci{+2.70}{+4.85} over the closed-loop focus map; on the 129 fast-motion videos it is $+6.89$ \ci{+4.03}{+9.49} over the closed-loop map.

\subsection{With a permanent-memory schedule}\label{sec:full}
Cutie writes working-memory frames and evicts old ones, and on long videos the object is lost for want of old, clean anchors. Writing every 20th frame to its permanent memory with Cutie's own prediction needs no training and no further annotation (Appendix~\ref{app:perm}; the method also runs that frame at full resolution). It is a property of the segmenter, so every encoder gains from it, and we report it as an additional operating point, not as part of the method. Table~\ref{tab:full} gives the held-out result with the same videos and heads as Table~\ref{tab:long}.

\begin{table}[htbp]
\centering
\small
\caption{Held-out LVOS V2 valid (140 videos, 239 objects; 42 in videos over 600 frames), three ViT-S heads, reference implementation (FLOPs frame-weighted across videos as everywhere; Table~\ref{tab:long} is the timed-implementation run, whose method row is 0.385). Schedule: every 20th frame written to Cutie's permanent memory. $\tau$ is the surprise threshold (0.06 in the method).}
\label{tab:full}
\begin{tabular}{llccc}
\toprule
Memory & Encoder & FLOPs & \jf{}, all & \jf{}, $>$600 frames \\
\midrule
Default & Dense & 1.000 & 69.47 & 52.27 \\
Default & ALGM 0.98 & 0.459 & 65.72 & 51.26 \\
Default & Ours & 0.388 & 68.63 & 51.99 \\
\midrule
Schedule & Dense & 1.000 & 72.88 & 62.26 \\
Schedule & ALGM 0.98 & 0.486 & 69.49 & 62.32 \\
Schedule & Ours & 0.413 & 72.82 & 63.74 \\
Schedule & Ours, $\tau=0.5$ & \textbf{0.335} & \textbf{73.61} & \textbf{65.33} \\
\bottomrule
\end{tabular}
\end{table}

With the schedule our method is level with dense Cutie given the same schedule ($-0.06$ \ci{-1.03}{+0.92}) at 0.413 of its FLOPs and $+3.34$ \ci{+1.91}{+4.94} over ALGM 0.98 at lower FLOPs (0.413 against 0.486). Raising the surprise threshold to 0.5 under the schedule lowers FLOPs to 0.335 (19\% fewer than ours with the schedule) at $+0.79$ \ci{-0.28}{+1.78} \jf{} over it, $+0.73$ \ci{-0.63}{+2.00} over dense with the schedule and $+4.13$ \ci{+2.68}{+5.79} over ALGM 0.98 with the schedule. Against dense Cutie with its default memory, the variant is $+4.14$ \ci{+2.49}{+5.77} at 0.335 of the FLOPs, and on the 42 objects in videos longer than 600 frames $+13.06$ \ci{+7.83}{+18.45}, or $+3.07$ \ci{-0.72}{+6.59} over dense with the schedule. We read this as the long-video result being at least as strong as the main one, under a memory that does not need training, and not as a better method: the variant was chosen on the development split, among five combinations of threshold and schedule, by a rule fixed in advance, and then run once on videos that had already been used to evaluate the method (Appendix~\ref{app:perm}). Its held-out acceptance criterion was written into the plan after its runs had finished and before any of its held-out numbers was computed, which is a weaker order than the primary protocol of Table~\ref{tab:long}, fixed before any run; the threshold 0.5 itself was proposed after its held-out ablation (Section~\ref{sec:ablation}) had been seen. Its intervals against dense with the schedule include zero, as the main claim's do. The schedule costs Cutie time that the encoder saving does not cover (a median step of 21.3 ms against 14.3 ms), so the FLOPs here are encoder FLOPs only, and the schedule raises peak GPU memory about sixfold on long videos (dense Cutie, 1.08 to 6.45 GiB; Section~\ref{sec:time}, Appendix~\ref{app:perm}).

\subsection{Where the gain comes from}\label{sec:attr}

\begin{table}[htbp]
\centering
\caption{Exact split of the long-video gain over ALGM 0.98 by frame class (head R31c, reference implementation; shares are of this head's scored object-frames). The classes follow the method's own state, so they show where the difference accrues, not what caused it.}
\label{tab:attr}
\begin{tabular}{lcc}
\toprule
Frames & Share & Contribution to the \jf{} gain \\
\midrule
Before the video's first fallback & 62.1\% & $\mathbf{+2.40}$ \ci{+0.68}{+4.20} \\
Fallback (every object lost; ALGM runs) & 19.6\% & $-1.13$ \ci{-2.04}{-0.44} \\
After re-acquisition & 18.2\% & $+1.56$ \ci{+0.65}{+2.64} \\
Total & & $+2.83$ \ci{+0.55}{+5.14} \\
\bottomrule
\end{tabular}
\end{table}

\begin{figure}[htbp]
\centering
\includegraphics[width=\linewidth]{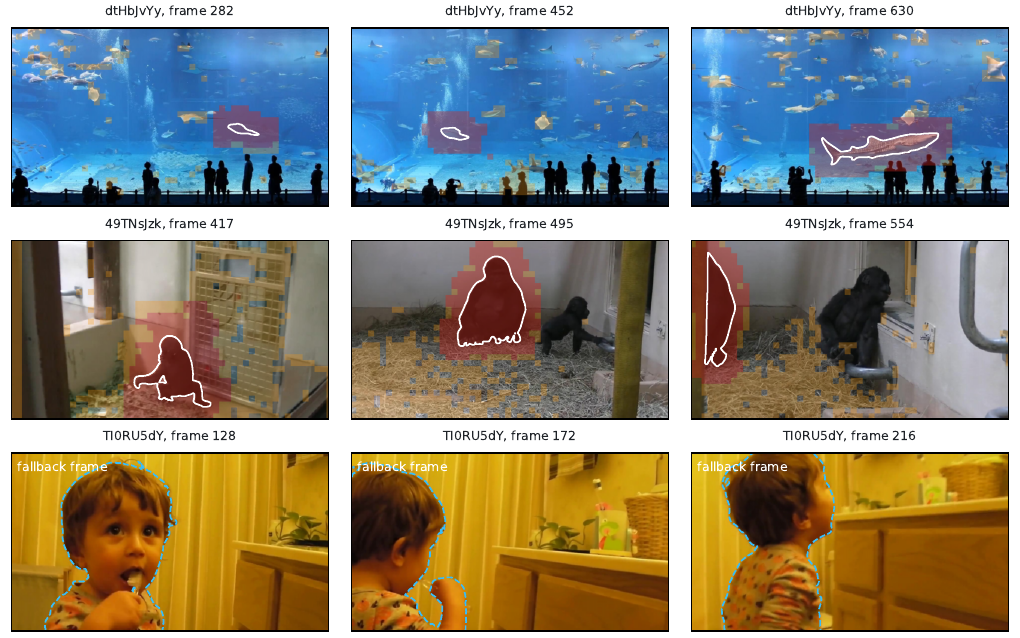}
\caption{Tiers and predictions on three held-out LVOS videos (head R31c). Rows 1 and 2: the method keeps a fast-moving target that ALGM 0.98 has lost (ALGM predicts no mask in these frames). Row 3: the video where the method does worst against ALGM; ALGM keeps the target (a child), the method has lost it. These frames already run the fallback (ALGM's own merging), so the loss happened earlier and persists through the memory. Red: high tier; orange: middle tier; white outline: the method's predicted mask; dashed blue: ALGM 0.98's. Frames are annotated frames where one of the two predicts no mask, whichever is more frequent in that video. Tiers replayed from the run's own predictions through the reference implementation, whose decisions the timed run equals.}
\label{fig:qual}
\end{figure}

Most of the gain accrues before any fallback (Table~\ref{tab:attr}); because these frame classes follow the method's own state, they describe where the gain accrues rather than isolating fallback or re-acquisition as causes. By outcome the gain is fewer object losses (defined here as ground-truth IoU below 0.1, distinct from the runtime controller's predicted-area rule in Section~\ref{sec:expect}): ALGM loses an object the method keeps on 7.0\% of frames, the reverse happens on 3.6\%, and on frames where both track the object they are level ($+0.36$ \ci{-0.24}{+0.87}). Figure~\ref{fig:qual} shows two videos where the method keeps a fast target that ALGM loses, and the video where it does worst.

\subsection{Do the safety nets cover the objects?}\label{sec:cover}
The premise is that the segmenter's previous mask misses fast objects and the expectation, the ring, the search region and the surprise check repair this. We measured it frame by frame (development split, 34 videos, one head, 36,450 object-frames; per frame the share of an object's ground-truth pixels inside each region). Table~\ref{tab:cover} gives the share inside the previous mask moved by the expected motion, inside the high tier the encoder finally ran, and inside the high or middle tier. For objects that move more than two patches per frame (17\% of object-frames) the moved mask misses a third of the frames (below 0.9 coverage on 33\%); the final high tier misses 2\%, and high or middle 19 of the 6,050 fast object-frames (0.31\%). Slow objects need no repair (1\% misses). The nets therefore do what they were built for. They do not make the segmenter find a lost target: in the first five frames after an object reappears (302 object-frames) the object lies 97\% inside the protected tiers, and on frames where it is visible but the prediction is empty (428 object-frames) 94\% inside them, yet the object's J (IoU) is 0.51 and 0 there (a descriptive count over the same frames, not a causal test). What limits long videos at that point is the segmenter's matching, not the encoder's coverage; Appendix~\ref{app:perm} shows a memory policy that acts on exactly this.

\begin{table}[htbp]
\centering
\small
\caption{Share of an object's ground truth inside each region (mean over object-frames; in brackets the share of frames below 0.9). Development split, 34 videos, the method at ViT-S.}
\label{tab:cover}
\begin{tabular}{lcccc}
\toprule
Object speed & Object-frames & Moved previous mask & Final high tier & High or middle \\
\midrule
slow ($<$0.5 patches) & 13,820 & 0.994 (1\%) & 0.996 (0\%) & 0.999 (0\%) \\
medium & 16,526 & 0.976 (3\%) & 0.987 (2\%) & 0.999 (0\%) \\
fast ($>$2 patches) & 6,050 & 0.851 (33\%) & 0.982 (2\%) & 0.999 (0.31\%) \\
all & 36,450 & 0.961 (7\%) & 0.989 (1\%) & 0.999 (0\%) \\
\bottomrule
\end{tabular}
\end{table}

\subsection{Ablations}\label{sec:ablation}
\begin{itemize}
\item \textbf{Merging the expected region costs accuracy.} Merging the high tier by ALGM's rules (everything else unchanged; reference implementation) lowers FLOPs by 10\% (0.348 against 0.388) and costs $-5.13$ \jf{} \ci{-6.91}{-3.48} on the held-out protocol, which puts it below plain ALGM ($-2.23$ \ci{-3.54}{-0.93}). The budget changes with it, so this shows that ALGM's merging there hurts, not that every protected patch is needed. On the development split, with an earlier surprise threshold, the same ablation costs $-2.47$ \ci{-3.50}{-1.56}.
\item \textbf{Against the closed-loop focus map.} Against the closed-loop map, which has no expectation or surprise check and falls back to ToMe r64 when its map is empty (so this compares the whole design, not the expectation alone), the method is $+3.79$ \jf{} on all 420 development videos and $+6.89$ on the fast-motion subset, at 0.377 against 0.268 FLOPs.
\item \textbf{Against the previous prediction under ALGM.} Keeping the closed-loop map at full resolution with plain ALGM 0.98 elsewhere and as the fallback, with no expectation or surprise check, matches the method's accuracy on the 420 development videos ($+1.36$ against $+1.33$ \jf{} over ALGM 0.98) but at 0.505 against 0.377 FLOPs. On the 140 held-out videos (three heads) it is $+3.45$ \ci{+2.01}{+5.03} over ALGM 0.98 and $+0.74$ \ci{-0.33}{+1.90} over the method, at 0.486 FLOPs against the method's 0.386 and ALGM's 0.459; on the 90 new videos it is ahead of the method ($+1.61$ \ci{+0.42}{+2.95}), and on the 42 objects in videos longer than 600 frames the two are level ($+0.17$ \ci{-2.44}{+2.63}). This control shows how much accuracy protecting the previous mask under ALGM already achieves at higher compute (0.486 FLOPs); the expectation and surprise tiers operate at lower counted encoder FLOPs (0.386) with a small accuracy trade-off, though this comparison cannot assign the gain to individual components.
\item \textbf{How much surprise is needed.} Raising the surprise threshold from 0.06 to 0.5 (reference implementation, held-out protocol, ablation after the main run) lowers FLOPs from 0.388 to 0.313 (19\%) at $-0.06$ \jf{} \ci{-1.10}{+1.00} against the method, and stays ahead of ALGM 0.98 ($+2.84$ \ci{+1.16}{+4.70}, against 0.459 FLOPs); turning the rule off gives $-1.10$ \ci{-2.38}{+0.12} at 0.307 FLOPs ($+1.80$ \ci{+0.04}{+3.68} over ALGM). The setting was chosen on the development split by a rule fixed in advance (it passed there: $+0.05$ at 0.269 against 0.366 FLOPs); on the held-out videos its lower bound is $-1.10$, short of the $-1.0$ the rule required, so it is reported as an ablation and not adopted. On the development split, dropping the ring as well costs $-1.73$ \ci{-3.77}{-0.15}: the ring and the surprise rule substitute for each other. The surprise check at its tuned threshold therefore buys little accuracy for about a fifth of the FLOPs.
\item \textbf{Headroom of the map's inputs.} Giving the method a perfect tracker's last two masks (the ground truth of the two previous frames, for the expectation and for the map; a diagnostic, not a method) on 104 development videos changes \jf{} by $+0.11$ \ci{-0.09}{+0.42} at 0.364 against 0.366 FLOPs ($+0.33$ \ci{-0.06}{+1.03} on the 44 long videos, $+0.40$ \ci{-0.09}{+1.31} on the fast-motion stratum). The map's inputs are therefore not what limits the method.
\item \textbf{Noise in the surprise check.} The variance floor and the 4-pixel tolerance halve false frame-wide surprises from sensor noise (12.8\% against 30.3\% of frames); camera compensation finer than one patch did not reduce them further.
\end{itemize}

\subsection{Short clips with every baseline}\label{sec:short}

\begin{table}[htbp]
\centering
\caption{DAVIS 2017 val inside Cutie, ViT-S (head R26; dense 87.26 \jf{}, our scorer), every evaluated setting (Expedite and PiToMe were run at ViT-B and ViT-L only); intervals over objects. Official scores and video-level intervals in Appendix~\ref{app:official}.}
\label{tab:davis_s}
\begin{tabular}{lcccc}
\toprule
Method & FLOPs & \jf{} & $-$ dense & Ours $-$ method \\
\midrule
Closed-loop focus map, rate 0.7 & 0.293 & 86.55 & $-0.71$ & $+0.51$ \ci{-0.17}{+1.69} \\
Protected ALGM 0.77 & 0.351 & 86.73 & $-0.53$ & $+0.33$ \ci{-0.01}{+0.91} \\
Closed-loop focus map, rate 0.3 & 0.353 & 86.54 & $-0.72$ & $+0.53$ \ci{+0.08}{+1.17} \\
Protected ALGM 0.83 & 0.373 & 86.74 & $-0.52$ & $+0.33$ \ci{+0.00}{+0.86} \\
ALGM 0.95 & 0.438 & 85.30 & $-1.96$ & $+1.77$ \ci{+0.89}{+2.52} \\
Protected ALGM 0.91 & 0.452 & 86.81 & $-0.45$ & $+0.26$ \ci{-0.03}{+0.75} \\
\textbf{Ours} & \textbf{0.462} & \textbf{87.07} & $\mathbf{-0.19}$ \ci{-0.75}{+0.18} & \\
AiluRus 0.39 & 0.550 & 87.41 & $+0.15$ & $-0.34$ \ci{-1.13}{+0.23} \\
ALGM 0.98 & 0.570 & 86.81 & $-0.45$ & $+0.26$ \ci{-0.44}{+0.76} \\
AiluRus 0.5 & 0.619 & 87.49 & $+0.23$ & $-0.42$ \ci{-1.11}{+0.10} \\
ToMe r80 & 0.650 & 87.43 & $+0.17$ & $-0.36$ \ci{-1.06}{+0.18} \\
ToMe r64 & 0.715 & 87.42 & $+0.16$ & $-0.35$ \ci{-1.11}{+0.22} \\
\bottomrule
\end{tabular}
\end{table}

\begin{table}[htbp]
\centering
\caption{DAVIS 2017 val, ViT-B (head R25; dense 88.44 \jf{}, our scorer), every evaluated setting.}
\label{tab:davis_b}
\begin{tabular}{lcccc}
\toprule
Method & FLOPs & \jf{} & $-$ dense & Ours $-$ method \\
\midrule
Closed-loop focus map, rate 0.7 & 0.378 & 87.68 & $-0.76$ & $+0.64$ \ci{-0.01}{+1.60} \\
Closed-loop focus map, rate 0.3 & 0.428 & 88.00 & $-0.44$ & $+0.32$ \ci{-0.14}{+1.11} \\
Protected ALGM 0.83 & 0.459 & 87.74 & $-0.70$ & $+0.58$ \ci{+0.09}{+1.45} \\
\textbf{Ours} & \textbf{0.524} & \textbf{88.32} & $\mathbf{-0.12}$ \ci{-0.27}{+0.01} & \\
ALGM 0.95 & 0.589 & 87.54 & $-0.90$ & $+0.78$ \ci{+0.47}{+1.17} \\
AiluRus 0.5 & 0.637 & 88.36 & $-0.08$ & $-0.04$ \ci{-0.22}{+0.12} \\
Protected ALGM 0.95 & 0.654 & 88.35 & $-0.09$ & $-0.03$ \ci{-0.13}{+0.07} \\
ALGM 0.97 & 0.655 & 88.09 & $-0.35$ & $+0.23$ \ci{+0.02}{+0.41} \\
ToMe r64 & 0.729 & 88.33 & $-0.11$ & $-0.01$ \ci{-0.15}{+0.11} \\
Expedite 0.49 & 0.769 & 86.09 & $-2.35$ & $+2.23$ \ci{+0.60}{+5.02} \\
PiToMe 0.97 & 0.818 & 87.98 & $-0.46$ & $+0.34$ \ci{+0.19}{+0.49} \\
\bottomrule
\end{tabular}
\end{table}

On short clips the method stays at dense level on both backbones (against a margin of 1 \jf{} below dense, chosen after the results, the lower bounds are $-0.75$ at ViT-S and $-0.27$ at ViT-B, so it is non-inferior; at ViT-L the bound, $-1.00$, sits at the margin) and is level with or ahead of every baseline it was paired with (Tables~\ref{tab:davis_s} and~\ref{tab:davis_b}). It is not the cheapest: the closed-loop focus map and protected ALGM reach within 1 \jf{} of dense at 0.29 to 0.46 FLOPs, while the task-agnostic mergers reach dense level only at 0.55 FLOPs or more. The method spends extra compute on short clips (rings, search regions, surprise) to stay robust on long videos; one method serves both. ViT-L rows are in Appendix~\ref{app:vitl}.

\subsection{Other backbones}\label{sec:backbones}
\textbf{ViT-B.} With the single ViT-B head R25 and the same held-out protocol against ALGM 0.95, the method is $+2.74$ \jf{} \ci{+0.85}{+4.91} at 0.430 against 0.488 FLOPs. With three ViT-B heads (R41 b1 to b3) on all 140 videos, the protocol of Table~\ref{tab:long} gives $+3.89$ \jf{} \ci{+2.03}{+5.87} at 0.434 against 0.488 FLOPs, superior under the pre-specified rule (official toolkit: $+3.95$ \ci{+2.10}{+5.94}); each head is above 0 ($+3.42$, $+4.04$, $+4.20$; resampling whole videos, $+3.89$ \ci{+1.38}{+6.52} pooled, and head b1's lower bound falls to $+0.01$). The gap to dense is larger than at ViT-S: $-3.20$ \ci{-4.79}{-1.61} (dense 72.82, ALGM 0.95 65.74, the method 69.62). The fallback runs on 32.9\% of frames.
\textbf{ViT-L, one head.} On all 140 videos the method is $+2.03$ \jf{} \ci{+0.12}{+4.06} against ALGM 0.95 at 0.436 against 0.548 FLOPs, and $-0.63$ \ci{-2.45}{+1.37} against dense (dense 74.64, ALGM 0.95 71.98, the method 74.01); official toolkit $+2.06$ \ci{+0.06}{+4.28}. With one head this is reported without a rule: the direction matches ViT-S and ViT-B, at 20\% fewer FLOPs than ALGM, and the method is level with dense; resampling whole videos the interval against ALGM includes 0 (Appendix~\ref{app:official}). On short clips it is $-0.15$ \ci{-1.00}{+0.46} against dense at 0.497 FLOPs (Appendix~\ref{app:vitl}).

\subsection{Wall clock}\label{sec:time}

\begin{table}[htbp]
\centering
\caption{Encoder time per frame (ms), batch 1, $448\times784$, bf16, idle 16 GB consumer GPU, median over 114 frames (three DAVIS clips, 38 frames each) whose maps and object states come from the ground-truth masks of the two previous frames, so no frame falls back; the encoder and the tier planning only, without Cutie's memory and decoder. ``Same implementation'': padded token buffers, CUDA graphs, compiled block segments, variable-length flash attention. Speed-up over dense in parentheses. Lower part: deterministic settings, as used for every accuracy number.}
\label{tab:time}
\begin{tabular}{lccc}
\toprule
Method & ViT-S & ViT-B & ViT-L \\
\midrule
Dense, CUDA graphs & 2.91 & 5.68 & 16.94 \\
ALGM, eager & 4.61 & 4.89 & -- \\
ALGM, same implementation & 1.60 & 3.18 & 9.46 \\
Ours, eager & 10.56 & 10.01 & -- \\
\textbf{Ours, same implementation} & \textbf{1.94} (1.50$\times$) & \textbf{3.46} (1.64$\times$) & \textbf{9.25} (1.83$\times$) \\
\midrule
Dense, deterministic & 3.33 & 6.16 & 18.07 \\
ALGM, deterministic & 1.83 & 3.42 & 9.69 \\
Ours, deterministic & 3.17 & 4.65 & 11.08 \\
\bottomrule
\end{tabular}
\end{table}

The timed implementation turns the FLOPs saving into a speed-up over dense on all three backbones (Table~\ref{tab:time}). ALGM given the same engineering is faster at ViT-S and ViT-B although it does more FLOPs (0.49 and 0.51 against 0.36 and 0.44 on these clips; 0.57 against 0.40 at ViT-L): our budget merges in every block and the expectation step take time that FLOPs do not count. A kernel profile of 60 frames places the difference in that bookkeeping: per frame the method launches about 820 (ViT-S) and 860 (ViT-B) GPU kernels against ALGM's 470, and its extra GPU time at ViT-S is in precision casts, gathers and scatters and small element-wise kernels (0.61 ms), not in attention or matrix products (0.07 ms). At ViT-L, where the encoder dominates, the method is as fast as ALGM (9.25 against 9.46 ms). With deterministic kernels the ViT-S speed-up over dense is small (1.05$\times$) and ALGM is faster than the method at every size. These clips start from ground-truth masks, so no frame falls back. Inside real long-video evaluations the gain shrinks. On 12 LVOS videos (6,516 frames, Cutie's own predicted masks, so the fallback runs on 39\% of frames at ViT-S and 49\% at ViT-B; an idle GPU) the encoder takes a median 4.20 ms against dense Cutie's 5.35 ms at ViT-S and 5.32 against 8.03 ms at ViT-B (means after each video's first 30 frames: 4.58 against 5.32 and 6.17 against 7.96), but the encoder is only 34\% to 44\% of Cutie's step, so the whole step falls by 3\% at ViT-S (15.18 against 15.63 ms) and 8\% at ViT-B (16.81 against 18.33 ms) in that steady state. The first frames of each video (graph capture and buffer sizing) cost about 1.4 s at ViT-S, so over videos of this length (543 frames on average) the mean step is 13\% above dense at ViT-S (17.65 against 15.57 ms) and 6\% above at ViT-B (19.30 against 18.27 ms), the medians are 5\% and 11\% below, and the 99th percentile of the encoder is about four times dense's (24.8 against 6.2 ms at ViT-S). On this 12-video sample, mean Cutie step time including startup was higher than for dense Cutie; the speed-up over dense holds for the encoder, in steady state, and grows with video length as the startup cost is spread. A rerun of the whole table after the fixes of Appendix~\ref{app:official} agrees with it within 1\% (4\% for deterministic ALGM at ViT-L). The threshold-0.5 setting of Section~\ref{sec:ablation} (an ablation that was not adopted, because its held-out lower bound missed the rule; 0.313 against 0.388 FLOPs in the ViT-S reference run with default memory) was timed in closed loop on the same 12 videos (not the scheduled variant of Section~\ref{sec:full}, which was not timed) and times a little lower: the encoder takes a median 3.94 ms at ViT-S and 5.09 ms at ViT-B (means after each video's first 30 frames 4.22 and 5.78 ms), and the whole step after startup 14.78 ms (5\% below dense) and 16.39 ms (11\% below). The 19\% reduction in counted encoder FLOPs from the ViT-S reference experiment thus corresponds to about 2 to 3 percentage points of steady-state step time in this separate timing measurement, because the encoder is a minority of the step. On the frames where the focus map itself runs (the non-fallback frames) the method's whole step after startup is 16.10 ms at ViT-S against dense Cutie's 15.63 ms and 18.37 against 18.33 ms at ViT-B, which is consistent with the steady-state saving of the main setting coming from the 39\% and 49\% of frames on which the fallback (plain ALGM) runs (the comparison is unmatched: dense Cutie's mean is over all frames and per-frame records were not kept); with the threshold-0.5 setting the non-fallback frames take 15.64 and 16.90 ms. Plain ALGM was also run in this closed-loop harness (Cutie's own masks, 12 videos), but only on the eager path: encoder median 7.70 ms against dense Cutie's 5.35 ms at ViT-S, and 7.68 against 8.03 ms at ViT-B. This is a different condition from Table~\ref{tab:time}, the equally engineered encoder benchmark with ground-truth histories, where ALGM is faster than the method at ViT-S and ViT-B; we did not engineer ALGM in closed loop, so the two conditions must not be mixed and the comparison with an equally engineered ALGM rests on Table~\ref{tab:time}.

\subsection{A second baseline: ToMe}\label{sec:tome}
We applied the same design with ToMe as the compared merger (ToMe's matching on the middle tier, plain ToMe r64 as the fallback). The first version is $-2.26$ \jf{} \ci{-3.93}{-0.71} against plain ToMe r64 at 0.449 against 0.709 FLOPs on the held-out protocol, so it is not non-inferior: ToMe merges a fixed 64 tokens per block, and once the budget has thinned the low tier those merges fall on the middle tier. A second version (ToMe's rule as a rate on the tiers) failed its development gate and was not validated. A third version adds two rules aimed at the failure modes found on development videos: a frame whose tracked object moves at the block-matching radius runs plain ToMe, and frames Cutie writes to memory run plain ToMe. It passed its gate and on the held-out protocol is non-inferior at 21\% lower FLOPs ($-0.68$ \ci{-1.56}{+0.10} at 0.557 against 0.709), not superior. Both numbers are from a rerun after a fix: the fallback frames of the first run let the focus map's interest leak into ToMe's merging (Appendix~\ref{app:official}), and the first-run values were $-2.23$ and $-0.45$. The held-out videos judged the ToMe arm twice, and both revisions were designed on the development split after the first failure; we report every attempt. With ToMe the design saves compute at ToMe's accuracy; with ALGM it also gains accuracy.

\section{Limitations}
\begin{itemize}
\item \textbf{Second baseline.} Against ToMe the design is non-inferior at lower FLOPs, not superior, and only after two revisions made after a held-out failure.
\item \textbf{Wall clock.} The encoder is faster than dense in steady state, but Cutie's whole step falls by only 3\% to 8\% in closed loop on LVOS and, with per-video startup, is not faster than dense on videos of about 540 frames (Section~\ref{sec:time}); at ViT-S and ViT-B the encoder is slower than ALGM's given the same engineering; with deterministic kernels the ViT-S speed-up over dense is small and ALGM is faster at every size. The ViT-L timed path has an unresolved intermittent crash, with the overlap on and off, that required relaunching long runs (Appendix~\ref{app:official}); the ViT-L timing (114 frames) completed with the overlap on.
\item \textbf{Fallback share.} On about 30\% of long-video frames, the policy switches to plain ALGM merging. Cutie continues from the memory accumulated on preceding frames.
\item \textbf{Per-split power and the gap to dense.} On LVOS V1 val alone (65 objects) the interval crosses 0, at ViT-S and at ViT-B; on the new V2 videos the method is below dense while above ALGM. At ViT-B the method stays 3.2 \jf{} below dense.
\item \textbf{Very long videos and lost targets.} The gain over ALGM comes from videos up to 600 frames (Table~\ref{tab:length}: $+3.60$ up to 300 frames, $+2.95$ from 301 to 600 at ViT-S, reference implementation); on the 42 objects in videos longer than 600 frames it is not significant. In the timed-implementation runs, which count the three 300-frame videos in the middle bin (300 to 600 frames inclusive), the middle bin gives $+4.35$ at ViT-S and $+4.83$ at ViT-B per object and the longer videos $+0.48$ (ViT-S), $+0.89$ (ViT-B) and $+0.32$ (ViT-L). Where the fallback runs on more than half of a video's frames, the method is level with ALGM ($+0.06$, $-0.36$, $-1.92$); these videos are grouped by an outcome of the method, so this shows where it helps, not why. Writing every 20th frame to Cutie's permanent memory lifts every encoder on these videos by 10 to 12 \jf{} (Appendix~\ref{app:perm}); it is a property of Cutie's memory, so it is reported as an additional operating point (Section~\ref{sec:full}) and not as part of the method. Keeping the gain on the longest videos is the next design step.
\item \textbf{Backbone trend on V1 val.} On the 50 V1 videos the gain over ALGM shrinks as the backbone grows ($+3.62$, $+0.88$, $-1.51$ at ViT-S, B, L; 65 objects, every interval crosses 0); on the 90 new videos it does not ($+2.43$, $+5.01$, $+3.36$).
\item \textbf{Motion model.} The camera estimate is one integer translation and block matching searches an axis-aligned window, so rotation, zoom, parallax, blur and repeated texture can mislead the expectation or the surprise check; the fallback changes the merger but cannot undo what earlier frames wrote to Cutie's memory.
\item \textbf{Scope.} One VOS model (Cutie), one backbone family (DINOv3), one head at ViT-L. The frozen-ViT Cutie is below the official Cutie on LVOS (official small 65.99 and base 69.86 on V1 val, 65 objects; our ViT-S head $-11.37$ \ci{-18.58}{-4.28} below the small model, our ViT-B head $-3.28$ \ci{-9.17}{+2.43} below the small and $-7.15$ \ci{-13.03}{-1.59} below the base model; earlier heads, one run), which lowers every method's absolute numbers alike. Its heads start from the official Cutie weights and so inherit their training data (Section~\ref{sec:setup}).
\item \textbf{Reuse of data.} The method and its gate subset were informed by diagnostics on the development split, and its second gate run reused that split after a failed first run. The held-out test ran once on the reference implementation with rules fixed before it; the timed implementation was then evaluated on the same videos four times, each rerun following an implementation fix found by comparing decisions with the reference path, never by accuracy, with the rules unchanged. All five runs are superior to ALGM 0.98 ($+2.63$ to $+2.90$ \jf), and all are reported (Appendix~\ref{app:impl}); a sixth, the rerun that saved masks for the official toolkit, gives $+2.71$ (Appendix~\ref{app:official}).
\item \textbf{Schedule operating points.} The rows with the permanent-memory schedule (Section~\ref{sec:full}) use a Cutie-level change that raises Cutie's step time (median $+49\%$, 90th percentile 17.5 to 47.9 ms) and peak GPU memory (1.08 to 6.45 GiB peak allocated over 12 videos, dense Cutie), neither of which the encoder FLOPs count. The threshold-0.5 variant was chosen on the development split among five combinations and run once on held-out videos that had already served the method; its intervals against dense Cutie with the same schedule include zero. The schedule was not run on a third, untouched split; the LVOS test server, which would provide one, was not used.
\item \textbf{Short clips.} On DAVIS the closed-loop focus map is cheaper at dense level; the method trades short-clip compute for long-video robustness.
\end{itemize}

\section{Conclusion}
A frozen DINOv3 ViT inside a trained Cutie head can spend its compute where the segmenter expects the objects to be. Under the pre-specified long-video protocol (LVOS V2 valid, three heads at ViT-S and ViT-B), full resolution in the expected region, the compared merger's own rules in an uncertain middle tier, and fallback when tracking fails beat ALGM at lower counted encoder FLOPs ($+2.76$ \jf{} at 16\% fewer FLOPs at ViT-S; $+3.89$ at ViT-B) and stay within $0.2$ \jf{} of the dense encoder on DAVIS 2017 val. The prior-mask ALGM control shows how much accuracy full resolution on the previous prediction already achieves at higher compute ($0.486$ FLOPs), while the expectation and surprise tiers operate below ALGM's FLOPs with a small accuracy trade-off. Separately, a training-free permanent-memory schedule for Cutie lifts every encoder on long videos (reaching $73.61$ \jf{} at $0.335$ encoder FLOPs with $\tau=0.5$, level with scheduled dense), at a $+49\%$ median step-time and $6\times$ GPU-memory cost (Section~\ref{sec:full}, Appendix~\ref{app:perm}). Finally, counted encoder FLOPs do not translate into wall-clock savings on our 12-video LVOS sample (mean startup-inclusive step time is above dense Cutie, and an equally engineered ALGM is faster at ViT-S and ViT-B; Section~\ref{sec:time}). Speed beyond an equally engineered ALGM at small backbones, and superiority over a second baseline, remain open.

\paragraph{Reproducibility.} The code, the ports of the baselines with their upstream commits (except DTEM, whose upstream repository has no licence, so the DTEM rows of Appendix~\ref{app:more} cannot be reproduced from the release), the evaluators, the analysis scripts and the summary result files are released at \url{https://github.com/boletoms/focused-cv} (Apache-2.0; one file derived from PiToMe keeps its CC BY-NC 4.0 licence). The rules for each adoption decision were written in dated plan files before the corresponding runs; the plan for this method is included in the release (\texttt{docs/R40\_expectation\_plan.md}).

\paragraph{Acknowledgments.} Some AI tool assistance was used for code review, verification checks, and copyediting. The authors take full responsibility for all experiments, results, citations, and text in this paper.

\bibliographystyle{plainnat}
\bibliography{refs}

\begin{thebibliography}{44}
\providecommand{\natexlab}[1]{#1}
\providecommand{\url}[1]{\texttt{#1}}
\expandafter\ifx\csname urlstyle\endcsname\relax
  \providecommand{\doi}[1]{doi: #1}\else
  \providecommand{\doi}{doi: \begingroup \urlstyle{rm}\Url}\fi

\bibitem[Bercy et~al.(2026)Bercy, Poreba, Szczepanski, and Bouchafa]{g2tm}
Victor Bercy, Martyna Poreba, Michal Szczepanski, and Samia Bouchafa.
\newblock Decoder-agnostic token merging for vision transformers: A systematic
  study of {G2TM}.
\newblock \emph{arXiv preprint arXiv:2609.18279}, 2026.

\bibitem[Bolya et~al.(2023)Bolya, Fu, Dai, Zhang, Feichtenhofer, and
  Hoffman]{bolya2023tome}
Daniel Bolya, Cheng-Yang Fu, Xiaoliang Dai, Peizhao Zhang, Christoph
  Feichtenhofer, and Judy Hoffman.
\newblock Token merging: Your {ViT} but faster.
\newblock In \emph{ICLR}, 2023.
\newblock arXiv:2210.09461.

\bibitem[Chen et~al.(2025)Chen, Ni, Huang, Liu, Jeong, Wen, Bastian, Latapie,
  and Imani]{vltp}
Hanning Chen, Yang Ni, Wenjun Huang, Yezi Liu, SungHeon Jeong, Fei Wen,
  Nathaniel Bastian, Hugo Latapie, and Mohsen Imani.
\newblock {VLTP}: Vision-language guided token pruning for task-oriented
  segmentation.
\newblock In \emph{WACV}, 2025.
\newblock arXiv:2409.08464.

\bibitem[Cheng and Schwing(2022)]{cheng2022xmem}
Ho~Kei Cheng and Alexander~G. Schwing.
\newblock {XMem}: Long-term video object segmentation with an
  {Atkinson-Shiffrin} memory model.
\newblock In \emph{ECCV}, 2022.
\newblock arXiv:2207.07115.

\bibitem[Cheng et~al.(2024)Cheng, Oh, Price, Lee, and Schwing]{cheng2024cutie}
Ho~Kei Cheng, Seoung~Wug Oh, Brian~L. Price, Joon-Young Lee, and Alexander~G.
  Schwing.
\newblock Putting the object back into video object segmentation.
\newblock In \emph{CVPR}, 2024.
\newblock arXiv:2310.12982.

\bibitem[Choi et~al.(2024)Choi, Lee, Chu, Choi, and Kim]{choi2024vidtldr}
Joonmyung Choi, Sanghyeok Lee, Jaewon Chu, Minhyuk Choi, and Hyunwoo~J. Kim.
\newblock {vid-TLDR}: Training free token merging for light-weight video
  transformer.
\newblock In \emph{CVPR}, 2024.
\newblock arXiv:2403.13347.

\bibitem[Choudhury et~al.(2024)Choudhury, Zhu, Liu, Niinuma, Kitani, and
  Jeni]{choudhury2024rlt}
Rohan Choudhury, Guanglei Zhu, Sihan Liu, Koichiro Niinuma, Kris~M. Kitani, and
  L{\'a}szl{\'o}~A. Jeni.
\newblock Don't look twice: Faster video transformers with run-length
  tokenization.
\newblock In \emph{NeurIPS}, 2024.
\newblock arXiv:2411.05222.

\bibitem[Dao(2024)]{dao2024flashattention2}
Tri Dao.
\newblock {FlashAttention-2}: Faster attention with better parallelism and work
  partitioning.
\newblock In \emph{ICLR}, 2024.
\newblock arXiv:2307.08691.

\bibitem[Ding et~al.(2023)Ding, Liu, He, Jiang, Torr, and Bai]{ding2023mose}
Henghui Ding, Chang Liu, Shuting He, Xudong Jiang, Philip H.~S. Torr, and Song
  Bai.
\newblock {MOSE}: A new dataset for video object segmentation in complex
  scenes.
\newblock In \emph{ICCV}, 2023.
\newblock arXiv:2302.01872.

\bibitem[Ding et~al.(2025)Ding, Qian, Dong, Zhang, Zang, Cao, Guo, Lin, and
  Wang]{ding2025sam2long}
Shuangrui Ding, Rui Qian, Xiaoyi Dong, Pan Zhang, Yuhang Zang, Yuhang Cao,
  Yuwei Guo, Dahua Lin, and Jiaqi Wang.
\newblock {SAM2Long}: Enhancing {SAM 2} for long video segmentation with a
  training-free memory tree.
\newblock In \emph{ICCV}, 2025.
\newblock arXiv:2410.16268.

\bibitem[Dutson et~al.(2023)Dutson, Li, and Gupta]{dutson2023eventful}
Matthew Dutson, Yin Li, and Mohit Gupta.
\newblock Eventful transformers: Leveraging temporal redundancy in vision
  transformers.
\newblock In \emph{ICCV}, 2023.
\newblock arXiv:2308.13494.

\bibitem[Dutta et~al.(2025)Dutta, Maity, and Mitra]{prato}
Pallabi Dutta, Anubhab Maity, and Sushmita Mitra.
\newblock Prompt-based dynamic token pruning for efficient segmentation of
  medical images.
\newblock \emph{arXiv preprint arXiv:2506.16369}, 2025.

\bibitem[Fu et~al.(2021)Fu, Zhao, Sun, Huang, Wang, and Ding]{fu2021motionroi}
Lihua Fu, Yu~Zhao, Xiaowei Sun, Jialiang Huang, Dan Wang, and Yu~Ding.
\newblock Video object segmentation based on motion-aware {ROI} prediction and
  adaptive reference updating.
\newblock \emph{Expert Systems with Applications}, 167:\penalty0 114153, 2021.

\bibitem[Glandorf et~al.(2026)Glandorf, Norrenbrock, and Rosenhahn]{vpp}
Patrick Glandorf, Thomas Norrenbrock, and Bodo Rosenhahn.
\newblock Video patch pruning: Efficient video instance segmentation via early
  token reduction.
\newblock In \emph{CVPR Workshops}, 2026.
\newblock arXiv:2604.00827.

\bibitem[Gong and Lis(2025)]{gong2025cubist}
Wenyi Gong and Mieszko Lis.
\newblock {CubistMerge}: Spatial-preserving token merging for diverse {ViT}
  backbones.
\newblock \emph{arXiv preprint arXiv:2509.21764}, 2025.

\bibitem[Holm(1979)]{holm1979}
Sture Holm.
\newblock A simple sequentially rejective multiple test procedure.
\newblock \emph{Scandinavian Journal of Statistics}, 6\penalty0 (2):\penalty0
  65--70, 1979.

\bibitem[Hong et~al.(2023)Hong, Chen, Liu, Zhang, Guo, Chen, and
  Zhang]{hong2023lvos}
Lingyi Hong, Wenchao Chen, Zhongying Liu, Wei Zhang, Pinxue Guo, Zhaoyu Chen,
  and Wenqiang Zhang.
\newblock {LVOS}: A benchmark for long-term video object segmentation.
\newblock In \emph{ICCV}, 2023.
\newblock arXiv:2211.10181.

\bibitem[Hong et~al.(2025)Hong, Liu, Chen, Tan, Feng, Zhou, Guo, Li, Chen, Gao,
  Zhang, and Zhang]{hong2025lvos2}
Lingyi Hong, Zhongying Liu, Wenchao Chen, Chenzhi Tan, Yuang Feng, Xinyu Zhou,
  Pinxue Guo, Jinglun Li, Zhaoyu Chen, Shuyong Gao, Wei Zhang, and Wenqiang
  Zhang.
\newblock {LVOS}: A benchmark for large-scale long-term video object
  segmentation.
\newblock \emph{IEEE Transactions on Pattern Analysis and Machine
  Intelligence}, 2025.
\newblock arXiv:2404.19326.

\bibitem[Kim and Lim(2026)]{sam2dual}
JeongRae Kim and Changwon Lim.
\newblock {SAM2Dual}: Training-free, dual memory for long-term video object
  segmentation.
\newblock \emph{arXiv preprint arXiv:2608.18640}, 2026.

\bibitem[Lee and Hong(2024)]{lee2024dtem}
Dong~Hoon Lee and Seunghoon Hong.
\newblock Learning to merge tokens via decoupled embedding for efficient vision
  transformers.
\newblock In \emph{NeurIPS}, 2024.
\newblock arXiv:2412.10569.

\bibitem[Li et~al.(2023)Li, Wang, Zhang, Shi, Jiang, Li, Dai, Xiong, and
  Tian]{li2023ailurus}
Jin Li, Yaoming Wang, Xiaopeng Zhang, Bowen Shi, Dongsheng Jiang, Chenglin Li,
  Wenrui Dai, Hongkai Xiong, and Qi~Tian.
\newblock {AiluRus}: A scalable {ViT} framework for dense prediction.
\newblock In \emph{NeurIPS}, 2023.
\newblock arXiv:2311.01197.

\bibitem[Liang et~al.(2022)Liang, Yuan, Ding, Luo, Lin, Jia, Zhang, Zhang, and
  Hu]{liang2022expedite}
Weicong Liang, Yuhui Yuan, Henghui Ding, Xiao Luo, Weihong Lin, Ding Jia, Zheng
  Zhang, Chao Zhang, and Han Hu.
\newblock Expediting large-scale vision transformer for dense prediction
  without fine-tuning.
\newblock In \emph{NeurIPS}, 2022.
\newblock arXiv:2210.01035.

\bibitem[Lu et~al.(2023)Lu, de~Geus, and Dubbelman]{lu2023cts}
Chenyang Lu, Daan de~Geus, and Gijs Dubbelman.
\newblock Content-aware token sharing for efficient semantic segmentation with
  vision transformers.
\newblock In \emph{CVPR}, 2023.
\newblock arXiv:2306.02095.

\bibitem[Mandal and Shashikumar(2026)]{rs3prune}
Avilasha Mandal and Sarvesh Shashikumar.
\newblock {RS$^3$-Prune}: Read-sparse, store-sparse token pruning for video
  object segmentation.
\newblock In \emph{BMVC}, 2026.
\newblock arXiv:2608.22526.

\bibitem[Mandal et~al.(2025)Mandal, Zhang, Puspitasari, Wang, Zhang, Qin, Wang,
  Yang, and Shen]{fastsam2}
Avilasha Mandal, Chaoning Zhang, Fachrina~Dewi Puspitasari, Xudong Wang,
  Jiaquan Zhang, Caiyan Qin, Guoqing Wang, Yang Yang, and Heng~Tao Shen.
\newblock Fast {SAM2} with text-driven token pruning.
\newblock \emph{arXiv preprint arXiv:2512.21333}, 2025.

\bibitem[Nguyen et~al.(2026)Nguyen, Tran, Nguyen, Xie, Nguyen, Truong,
  Palenicek, Le, Barz, Hannus, Nguyen, Dam, Le, Vu, Doan, Ngo, Xie, Zou,
  Sonntag, Peters, and Niepert]{structsam}
Duy M.~H. Nguyen, Tuan~A. Tran, Duong Nguyen, Siwei Xie, Trung~Q. Nguyen, Mai
  T.~N. Truong, Daniel Palenicek, An~T. Le, Michael Barz, Eric Hannus, TrungTin
  Nguyen, Tuan Dam, Ngan Le, Minh Vu, Khoa~D. Doan, Vien Ngo, Pengtao Xie,
  James Zou, Daniel Sonntag, Jan Peters, and Mathias Niepert.
\newblock {StructSAM}: Structure- and spectrum-preserving token merging for
  segment anything models.
\newblock \emph{arXiv preprint arXiv:2603.07307}, 2026.

\bibitem[Norouzi et~al.(2024)Norouzi, Orlova, de~Geus, and
  Dubbelman]{norouzi2024algm}
Narges Norouzi, Svetlana Orlova, Daan de~Geus, and Gijs Dubbelman.
\newblock {ALGM}: Adaptive local-then-global token merging for efficient
  semantic segmentation with plain vision transformers.
\newblock In \emph{CVPR}, 2024.
\newblock arXiv:2406.09936.

\bibitem[Pollard and Wray(2025)]{pollard2025video}
Sam Pollard and Michael Wray.
\newblock Video, how do your tokens merge?
\newblock In \emph{CVPR Workshop on Efficient Large Vision Models (eLVM)},
  2025.

\bibitem[Pont-Tuset et~al.(2017)Pont-Tuset, Perazzi, Caelles, Arbel{\'a}ez,
  Sorkine-Hornung, and Gool]{pont2017davis}
Jordi Pont-Tuset, Federico Perazzi, Sergi Caelles, Pablo Arbel{\'a}ez,
  Alexander Sorkine-Hornung, and Luc~Van Gool.
\newblock The 2017 {DAVIS} challenge on video object segmentation.
\newblock \emph{arXiv preprint arXiv:1704.00675}, 2017.

\bibitem[Rav{\'e} et~al.(2026)Rav{\'e}, Rasti, and Rousseau]{rave2026mpm}
Simon Rav{\'e}, Pejman Rasti, and David Rousseau.
\newblock {MPM}: Mutual pair merging for efficient vision transformers.
\newblock In \emph{CVPR Findings}, 2026.
\newblock arXiv:2604.05718.

\bibitem[Ravi et~al.(2025)Ravi, Gabeur, Hu, Hu, Ryali, Ma, Khedr, R{\"a}dle,
  Rolland, Gustafson, Mintun, Pan, Alwala, Carion, Wu, Girshick, Doll{\'a}r,
  and Feichtenhofer]{ravi2025sam2}
Nikhila Ravi, Valentin Gabeur, Yuan-Ting Hu, Ronghang Hu, Chaitanya Ryali,
  Tengyu Ma, Haitham Khedr, Roman R{\"a}dle, Chloe Rolland, Laura Gustafson,
  Eric Mintun, Junting Pan, Kalyan~Vasudev Alwala, Nicolas Carion, Chao-Yuan
  Wu, Ross Girshick, Piotr Doll{\'a}r, and Christoph Feichtenhofer.
\newblock {SAM 2}: Segment anything in images and videos.
\newblock In \emph{ICLR}, 2025.
\newblock arXiv:2408.00714.

\bibitem[Ren et~al.(2026)Ren, Gong, and Zeng]{scmerge}
Junfeng Ren, Yi~Gong, and Yuan Zeng.
\newblock {SCMerge}: Spatial-consistency-aware token pair merging for efficient
  semantic segmentation.
\newblock \emph{arXiv preprint arXiv:2610.21642}, 2026.

\bibitem[Sarkar et~al.(2025)Sarkar, Datta, Kundu, Zheng, Bhattacharyya, and
  Beerel]{maskvd}
Sreetama Sarkar, Gourav Datta, Souvik Kundu, Kai Zheng, Chirayata
  Bhattacharyya, and Peter~A. Beerel.
\newblock {MaskVD}: Region masking for efficient video object detection.
\newblock In \emph{WACV}, 2025.
\newblock arXiv:2407.12067.

\bibitem[Schmidt and Newcombe(2025)]{schmidt2025stt}
Tanner Schmidt and Richard Newcombe.
\newblock Segment this thing: Foveated tokenization for efficient
  point-prompted segmentation.
\newblock In \emph{CVPR}, 2025.
\newblock arXiv:2506.11131.

\bibitem[Sheng et~al.(2026)Sheng, Liu, and Chen]{sheng2026codec}
Xihua Sheng, Ruijie Liu, and Chang~Wen Chen.
\newblock Video codec-guided training-free token compression for video large
  language models.
\newblock \emph{arXiv preprint arXiv:2610.12959}, 2026.

\bibitem[Sim{\'e}oni et~al.(2025)Sim{\'e}oni, Vo, Seitzer, Baldassarre, Oquab,
  Jose, Khalidov, Szafraniec, Yi, Ramamonjisoa, Massa, Haziza, Wehrstedt, Wang,
  Darcet, Moutakanni, Sentana, Roberts, Vedaldi, Tolan, Brandt, Couprie,
  Mairal, J{\'e}gou, Labatut, and Bojanowski]{simeoni2025dinov3}
Oriane Sim{\'e}oni, Huy~V. Vo, Maximilian Seitzer, Federico Baldassarre, Maxime
  Oquab, Cijo Jose, Vasil Khalidov, Marc Szafraniec, Seungeun Yi, Micha{\"e}l
  Ramamonjisoa, Francisco Massa, Daniel Haziza, Luca Wehrstedt, Jianyuan Wang,
  Timoth{\'e}e Darcet, Th{\'e}o Moutakanni, Leonel Sentana, Claire Roberts,
  Andrea Vedaldi, Jamie Tolan, John Brandt, Camille Couprie, Julien Mairal,
  Herv{\'e} J{\'e}gou, Patrick Labatut, and Piotr Bojanowski.
\newblock {DINOv3}.
\newblock \emph{arXiv preprint arXiv:2508.10104}, 2025.

\bibitem[Tang et~al.(2023)Tang, Zhang, Liu, Liu, and Liu]{tang2023dtop}
Quan Tang, Bowen Zhang, Jiajun Liu, Fagui Liu, and Yifan Liu.
\newblock Dynamic token pruning in plain vision transformers for semantic
  segmentation.
\newblock In \emph{ICCV}, 2023.
\newblock arXiv:2308.01045.

\bibitem[Tang et~al.(2026)Tang, Liang, Wan, Zhu, Xu, Zhou, Song, You, and
  Zhao]{tang2026rds}
Weidong Tang, Zhiyuan Liang, Xinyan Wan, Chen Zhu, Zhaopan Xu, Pengfei Zhou,
  Yan Song, Yang You, and Wangbo Zhao.
\newblock Efficient video object segmentation and tracking with recurrent
  dynamic submodel.
\newblock In \emph{CVPR}, 2026.

\bibitem[Tran et~al.(2024)Tran, Nguyen, Nguyen, Le, Xie, Sonntag, Zou, Nguyen,
  and Niepert]{tran2024pitome}
Hoai-Chau Tran, Duy M.~H. Nguyen, Trung-Tin Nguyen, Ngan Le, Pengtao Xie,
  Daniel Sonntag, James~Y. Zou, Binh~T. Nguyen, and Mathias Niepert.
\newblock Accelerating transformers with spectrum-preserving token merging.
\newblock In \emph{NeurIPS}, 2024.
\newblock arXiv:2405.16148.

\bibitem[Xiong et~al.(2025)Xiong, Zhou, Xiang, Wu, Zhu, Liu, Suri, Varadarajan,
  Akula, Iandola, Krishnamoorthi, Soran, and Chandra]{xiong2024efficienttam}
Yunyang Xiong, Chong Zhou, Xiaoyu Xiang, Lemeng Wu, Chenchen Zhu, Zechun Liu,
  Saksham Suri, Balakrishnan Varadarajan, Ramya Akula, Forrest Iandola,
  Raghuraman Krishnamoorthi, Bilge Soran, and Vikas Chandra.
\newblock {EfficientTAM}: Track anything with lightweight vision transformers.
\newblock In \emph{ICCV}, 2025.
\newblock arXiv:2411.18933.

\bibitem[Yang et~al.(2024)Yang, Huang, Chai, Jiang, and Hwang]{yang2024samurai}
Cheng-Yen Yang, Hsiang-Wei Huang, Wenhao Chai, Zhongyu Jiang, and Jenq-Neng
  Hwang.
\newblock {SAMURAI}: Adapting segment anything model for zero-shot visual
  tracking with motion-aware memory.
\newblock \emph{arXiv preprint arXiv:2411.11922}, 2024.

\bibitem[Yung et~al.(2026)Yung, Bragman, Xu, Luengo, Stoyanov, and
  Mazomenos]{tca}
Ka-Wai Yung, Felix J.~S. Bragman, Jialang Xu, Imanol Luengo, Danail Stoyanov,
  and Evangelos~B. Mazomenos.
\newblock Temporal cluster assignment for efficient real-time video
  segmentation.
\newblock In \emph{CVPR Workshops}, 2026.
\newblock arXiv:2508.05851.

\bibitem[Zhang et~al.(2026)Zhang, Li, Liu, and Gu]{efficientsam2}
Jing Zhang, Zhikai Li, Xuewen Liu, and Qingyi Gu.
\newblock {Efficient-SAM2}: Accelerating {SAM2} with object-aware visual
  encoding and memory retrieval.
\newblock In \emph{ICLR}, 2026.
\newblock arXiv:2602.08224.

\bibitem[Zhou et~al.(2025)Zhou, Zhu, Xiong, Suri, Xiao, Wu, Krishnamoorthi,
  Dai, Loy, Chandra, and Soran]{zhou2025edgetam}
Chong Zhou, Chenchen Zhu, Yunyang Xiong, Saksham Suri, Fanyi Xiao, Lemeng Wu,
  Raghuraman Krishnamoorthi, Bo~Dai, Chen~Change Loy, Vikas Chandra, and Bilge
  Soran.
\newblock {EdgeTAM}: On-device track anything model.
\newblock In \emph{CVPR}, 2025.
\newblock arXiv:2501.07256.

\end{thebibliography}

\appendix

\section{Implementation equivalence and every held-out run}\label{app:impl}
Three implementation differences were found and fixed before the reported rows, each by checking that the timed implementation took the reference path's decisions: compiled kernels kept fp32 intermediates where the reference path rounds to bf16 (fixed by emulating the casts); ALGM's threshold statistics were built in fp32 where the reference path keeps them in bf16 (fixed); and a lost object's search region grew to the whole frame after a loss longer than the frame's width plus height in patches, which only long videos contain (fixed). A replay of the ground-truth masks of all 140 long videos through both implementations, at the ViT-S and ViT-B settings, agrees on every fallback decision and on the tiers of all but one of 61,880 frames. On fallback frames the timed implementation does not compute the expectation's products, which no later frame reads, and its counted FLOPs exclude them.

All five runs on the held-out videos against ALGM 0.98, in order: reference implementation $+2.90$ \ci{+1.28}{+4.67} at 0.388; timed implementation $+2.64$ \ci{+0.92}{+4.32} at 0.401; after the first fix $+2.82$ \ci{+1.24}{+4.49} at 0.395; after the second $+2.63$ \ci{+1.13}{+4.26} at 0.397; after the third (reported) $+2.76$ \ci{+1.22}{+4.44} at 0.385. Per head the reference and the reported timed rows differ by $+0.33$, $-0.33$ and $-0.45$ \jf, all intervals containing 0. Table~\ref{tab:attr} and the ablations use the reference implementation.

\section{Official scores and reruns}\label{app:official}
The official toolkits need saved masks, which the reported runs did not keep, so every reported row was rerun with masks saved: LVOS with the LVOS evaluation code~\cite{hong2023lvos} (commit 25cb02d), DAVIS with the DAVIS 2017 evaluation code (commit ac7c43f). Ground-truth masks score 100 under both. Our scorer averages every annotated frame per object; the LVOS toolkit scores every fifth frame of an object's range without its first and last, and the DAVIS toolkit leaves out the first and last frame.

\textbf{Reruns.} Every eager row (dense and each baseline, long videos and short clips) reproduces the reported rows exactly, object by object. The method's timed implementation does not: on the long videos 657 of 717 object rows (three heads) are equal; of the other 60, 58 differ by less than 1 \jf{}, one by 1.3, and one by 32 after a loss in one run and not in the other. On the rerun the pre-specified difference against ALGM 0.98 is $+2.71$ \ci{+1.20}{+4.49} on our scorer (reported: $+2.76$ \ci{+1.22}{+4.44}), and the method is superior under either. The short-clip rows of the method reproduce the reported \jf{}.

\textbf{ToMe fallback fix.} An audit found that on fallback frames (no object map) the ToMe variants of the method still read the previous interest map, so ToMe's merging was not the plain baseline's on those frames. The fix makes the fallback frame run plain ToMe; the ToMe rows of Section~\ref{sec:tome} were rerun on the held-out videos with all three heads (first and third version; the method with ALGM is unaffected, as it never reads the map on fallback frames), and the corrected values are the ones reported there.

\textbf{Long videos (ViT-S, three heads; rerun).} Official \jf{}: dense 69.19, ALGM 0.98 65.40, the method 68.16. Ours minus ALGM 0.98 under the official toolkit: $+2.75$ \ci{+1.23}{+4.55} on all 239 objects, $+3.64$ \ci{-0.26}{+7.27} on LVOS V1 val (65), $+2.42$ \ci{+0.66}{+4.13} on the 90 new videos (174); per head $+3.07$ \ci{+0.83}{+5.42}, $+2.74$ \ci{+0.56}{+5.17} and $+2.45$ \ci{-0.23}{+5.39}. Ours minus dense: $-1.03$ \ci{-2.35}{+0.25}. The toolkit's own summary (mean of seen and unseen J and F) per head, dense / ALGM 0.98 / ours: 70.18 / 66.02 / 69.87, 69.10 / 66.15 / 69.91 and 66.96 / 61.01 / 62.99. Holm-adjusted p-values of the secondary comparisons against ALGM 0.98 (our scorer, rerun; whole videos resampled, 10,000 draws): per head 0.08, 0.11 and 0.16; LVOS V1 val 0.16; the 90 new videos 0.07, so none is significant at 0.05 after correction.

\textbf{Short clips.} Official \jf{} of dense: 86.60 (ViT-S), 87.95 (ViT-B); the method: 86.40 and 87.81, i.e.\ $-0.19$ \ci{-0.78}{+0.20} and $-0.14$ \ci{-0.31}{+0.01} against dense. Against the baselines of Tables~\ref{tab:davis_s} and~\ref{tab:davis_b}, the Holm-adjusted differences (whole videos resampled) that remain significant at 0.05 under both scorers are those against ALGM 0.95 at ViT-S and ViT-B and against PiToMe at ViT-B (all in the method's favour, $p\le0.006$), and, at ViT-L, those against Expedite and ToMe r64 ($p\le0.003$); Expedite at ViT-B is not ($p=0.51$ and $0.47$), and every other difference has an adjusted $p\ge0.05$ under at least one scorer.

\textbf{Long videos (ViT-B, three heads).} Official \jf{}: dense 72.57, ALGM 0.95 65.44, the method 69.39. Ours minus ALGM 0.95 under the official toolkit: $+3.95$ \ci{+2.10}{+5.94} on all 239 objects, $+0.91$ \ci{-2.95}{+4.90} on LVOS V1 val, $+5.09$ \ci{+3.01}{+7.42} on the 90 new videos; per head $+3.48$ \ci{+1.08}{+6.13}, $+4.10$ \ci{+1.55}{+6.85} and $+4.28$ \ci{+2.19}{+6.60}. Ours minus dense: $-3.17$ \ci{-4.75}{-1.59}. The toolkit's own summary per head, dense / ALGM 0.95 / ours: 71.05 / 62.38 / 67.96, 72.34 / 63.37 / 69.61 and 72.16 / 62.81 / 67.14. Holm-adjusted p-values of the secondary comparisons (our scorer; whole videos resampled): per head 0.10, 0.03 and 0.03; LVOS V1 val 0.66; the 90 new videos 0.007.

\textbf{Whole videos resampled.} Objects of one video share frames and tracker state. Resampling the 140 videos (10,000 draws within the V1 and the new strata) instead of objects, on the reported rows: ViT-S against ALGM 0.98 $+2.76$ \ci{+0.99}{+4.52}, with videos weighted equally $+2.81$ \ci{+0.73}{+5.00}; on the 90 new videos $+2.43$ \ci{+0.54}{+4.25} and $+1.65$ \ci{-0.33}{+3.69}; against dense $-1.00$ \ci{-2.16}{+0.19} (90 new videos: $-1.95$ \ci{-3.24}{-0.73}). ViT-B against ALGM 0.95 $+3.89$ \ci{+1.38}{+6.52}, equally weighted $+3.11$ \ci{+0.94}{+5.41}; against dense $-3.20$ \ci{-4.99}{-1.57}. ViT-L against ALGM 0.95 $+2.03$ \ci{-0.20}{+4.33}, equally weighted $+2.37$ \ci{+0.07}{+4.93}; on the 90 new videos $+3.36$ \ci{+0.98}{+5.93} (Holm-adjusted $p=0.026$); against dense $-0.63$ \ci{-2.83}{+1.56}. At ViT-L, with one head, the overall difference against ALGM is therefore above 0 when objects are resampled but not when videos are (unadjusted $p=0.077$). The pre-specified test and its conclusion are unchanged: both ALGM comparisons stay above 0 (\texttt{r40\_video\_cluster.json}).

\textbf{Long videos (ViT-L, one head).} Official \jf{}: dense 74.42, ALGM 0.95 71.73, the method 73.79; ours minus ALGM 0.95 $+2.06$ \ci{+0.06}{+4.28} (LVOS V1 val $-1.53$ \ci{-6.11}{+2.74}, the 90 new videos $+3.41$ \ci{+1.27}{+5.76}); ours minus dense $-0.63$ \ci{-2.62}{+1.50}. Toolkit summary, dense / ALGM 0.95 / ours: 74.82 / 70.71 / 73.55. On DAVIS the method is $-0.14$ \ci{-0.96}{+0.44} against dense (official 87.98 against 88.13).

\textbf{ViT-L runs.} The method's ViT-L timed path crashes intermittently, with an illegal memory access in an encoder graph replay a few seconds into a new video, with and without MPS and with the side-stream overlap (block 0's attention beside the planning step) on and off. On a fixed 13-video sequence the crash reproduces at the 13th video and disappears with the overlap off or when the tracker's stale per-state graphs are never destroyed, but on the full run neither change prevents it (a run with the overlap off crashed after 10 videos, and so did one that kept the stale graphs). It persists with synchronous kernel launches, with the side graph replayed on the main stream, with separate graph pools, with the allocator's expandable segments off and with the device synchronized before each destroy, so it is not a stream race. The cause is open. The reported ViT-L rows come from one mode: overlap off, one process, no MPS, relaunched after each crash (rows are written per video, and the video a crash interrupted reruns from its start): three crashes in the first 42 videos, none in the remaining 98 videos and the 30 short clips. The overlap only reorders work; a rerun of three videos saved with the overlap on gave identical rows for two and object differences of $+0.6$ and $+1.7$ J for the third, within the timed implementation's rerun variation. ViT-S and ViT-B rows and timings ran with the overlap and never crashed.

\section{ViT-L}\label{app:vitl}
Table~\ref{tab:davis_l} gives DAVIS 2017 val at ViT-L (head R34) with every baseline; the method's row ran with the overlap off (Appendix~\ref{app:official}). The method is at dense level and level with or ahead of every baseline. ToMe r64 merges 64 tokens in each of 24 blocks and falls 10 \jf{} below dense. Long videos (one head, all 140 videos, 239 objects): the method 74.01 \jf{} at 0.436 FLOPs, ALGM 0.95 71.98 at 0.548, dense 74.64; ours minus ALGM 0.95 $+2.03$ \ci{+0.12}{+4.06}, $-1.51$ \ci{-6.12}{+2.76} on LVOS V1 val and $+3.36$ \ci{+1.21}{+5.73} on the 90 new videos; ours minus dense $-0.63$ \ci{-2.45}{+1.37}; ALGM 0.95 minus dense $-2.66$ \ci{-4.85}{-0.62}. The fallback runs on 30.0\% of frames.

\begin{table}[htbp]
\centering
\small
\caption{DAVIS 2017 val, ViT-L (head R34; dense 88.60 \jf{}, our scorer).}
\label{tab:davis_l}
\begin{tabular}{lcccc}
\toprule
Method & FLOPs & \jf{} & $-$ dense & Ours $-$ method \\
\midrule
Closed-loop focus map, rate 0.7 & 0.303 & 88.49 & $-0.11$ & $-0.04$ \ci{-1.05}{+0.93} \\
Closed-loop focus map, rate 0.3 & 0.330 & 88.12 & $-0.49$ & $+0.33$ \ci{-0.01}{+0.74} \\
ToMe r64 & 0.495 & 78.56 & $-10.04$ & $+9.89$ \ci{+6.77}{+13.98} \\
\textbf{Ours} & \textbf{0.497} & \textbf{88.45} & $\mathbf{-0.15}$ \ci{-1.00}{+0.46} & \\
AiluRus 0.5 & 0.569 & 87.97 & $-0.63$ & $+0.48$ \ci{+0.17}{+0.84} \\
Protected ALGM 0.83 & 0.581 & 88.47 & $-0.13$ & $-0.03$ \ci{-0.86}{+0.60} \\
ALGM 0.95 & 0.650 & 88.23 & $-0.38$ & $+0.22$ \ci{-0.57}{+0.85} \\
ALGM 0.97 & 0.680 & 88.34 & $-0.26$ & $+0.11$ \ci{-0.64}{+0.69} \\
PiToMe 0.97 & 0.680 & 88.04 & $-0.56$ & $+0.41$ \ci{-0.47}{+1.11} \\
Expedite 0.49 & 0.696 & 87.13 & $-1.47$ & $+1.32$ \ci{+0.58}{+1.87} \\
Protected ALGM 0.95 & 0.700 & 88.48 & $-0.12$ & $-0.03$ \ci{-0.89}{+0.61} \\
\bottomrule
\end{tabular}
\end{table}

\section{A permanent-memory schedule for Cutie}\label{app:perm}
This appendix reports a finding about Cutie's memory, not about the focus map, and the controls that show it. Cutie writes a working-memory frame every fifth frame and evicts old ones; masks given by the user are permanent. On long videos the object is lost for want of old, clean anchors. The schedule writes every 20th frame, with Cutie's own predicted mask, to the permanent memory (the method also runs that frame at full resolution, as an I-frame; dense Cutie runs every frame at full resolution anyway). Periodic and permanent memories are not new in themselves: XMem~\cite{cheng2022xmem} and Cutie keep the given annotation permanently, and SAM2Dual~\cite{sam2dual} fills a long-term memory by interval sampling; what we report is the effect of a 20-frame permanent write of Cutie's own predictions, with controls on three encoders. No training; each object receives its standard first-appearance annotation, and scheduled writes store the model's own predictions. It was chosen on the development split (420 LVOS V2 train videos, one head) from seven frame schedules (a dense or fallback frame every 10, 20 or 30 frames, or on loss; none passed its pre-specified gate alone) and three memory treatments (forced write, clearing then writing, permanent write; only the permanent write passed), and then run once on the held-out videos with all three heads. Because the held-out gain was far larger than the development gain and above dense, controls were added: plain ALGM 0.98 and dense Cutie with the same schedule on the same videos.

\begin{table}[htbp]
\centering
\small
\caption{Permanent-memory schedule, held-out LVOS V2 valid (140 videos, 239 objects; 42 objects in the 600-frame-plus videos), three ViT-S heads, reference implementation. Gain: schedule minus default, per object, 95\% bootstrap interval.}
\label{tab:perm}
\begin{tabular}{lcccccc}
\toprule
 & \multicolumn{2}{c}{Default \jf{}} & \multicolumn{2}{c}{Schedule \jf{}} & \multicolumn{2}{c}{Gain, all objects} \\
 & all & $>$600 & all & $>$600 & & FLOPs \\
\midrule
Dense Cutie & 69.47 & 52.27 & 72.88 & 62.26 & $+3.41$ \ci{+2.31}{+4.51} & 1.000 \\
ALGM 0.98 & 65.72 & 51.26 & 69.49 & 62.32 & $+3.77$ \ci{+2.40}{+5.14} & 0.459 $\to$ 0.486 \\
\textbf{Ours} & 68.63 & 51.99 & 72.82 & 63.74 & $+4.20$ \ci{+2.57}{+5.74} & 0.388 $\to$ 0.413 \\
\bottomrule
\end{tabular}
\end{table}

Every setting gains, about 3.4 to 4.2 \jf{} overall and 10 to 12 on the objects in videos longer than 600 frames, so the gain belongs to Cutie's memory and is available to any encoder; dense Cutie with the schedule is the new dense bar. At equal schedule our method is $+3.34$ \ci{+1.91}{+4.94} over ALGM 0.98 at 0.413 against 0.486 FLOPs ($+1.42$ \ci{-0.95}{+4.21} on the 600-frame-plus videos, $+3.23$ \ci{+1.48}{+5.13} on the 90 new videos), and $-0.06$ \ci{-1.03}{+0.92} against dense with the schedule ($+1.48$ \ci{-1.65}{+4.37} and $-0.75$ \ci{-1.75}{+0.23}). The paper's claim therefore holds under both memory regimes. On the 316 development videos not used to choose the schedule (one head, R26; all at most 600 frames, since the 44 longer ones were in the selection set) the gains are smaller and in the same direction: dense Cutie $+0.71$ \ci{+0.11}{+1.36}, ours $+1.13$ \ci{+0.40}{+1.89}, ALGM 0.98 $+1.33$ \ci{+0.53}{+2.29} (454 objects); at equal schedule ours is $+0.96$ \ci{+0.31}{+1.73} over ALGM and $-0.85$ \ci{-1.66}{-0.16} against dense with the schedule, so on shorter videos dense with the schedule is ahead of ours. The effect grows with video length, and the 600-frame-plus videos are covered only by the held-out run. We keep the schedule out of the method (Section~\ref{sec:full} reports it as an operating point): it is a property of the segmenter, it was selected with the held-out videos of the method still unused for it only in part (they had served the method before), and it needs fresh data and an end-to-end cost before adoption. The encoder FLOPs do not count Cutie's own cost, and it is not small: the permanent memory grows by one frame per 20 frames, and over 12 long videos (6,516 frames, dense encoder, ViT-S, an otherwise idle GPU, one process per setting) the median time of Cutie's whole step per frame rises from 14.3 to 21.3 ms ($+49$\%; mean 15.2 to 26.2 ms, $+73$\%; 90th percentile 17.5 to 47.9 ms, 99th 31.0 to 62.1 ms) while the encoder stays at 5.0 ms (an earlier run of the same videos gave a median 14.9 to 21.8 ms, $+47$\%). The schedule's cost therefore exceeds the encoder time that any token reduction can save (5.0 ms of a 14.3 ms step), and it grows with the length of the video. Peak GPU memory of dense Cutie for the whole process rises from 1.08 GiB without the schedule to 6.45 GiB with it (peak allocated; reserved 1.25 to 6.82 GiB); the schedule's permanent memory is a real cost, not a rounding.

\section{Two further baselines and the gain by video length}\label{app:more}
\textbf{G2TM and DTEM on short clips.} G2TM~\cite{g2tm} is a recent training-free graph-based merger for segmentation (authors' code, frozen DINOv3 ViT-S, no fine-tuning); DTEM~\cite{lee2024dtem} learns its merging metric, and we trained only those layers through Cutie's own loss with the ViT and the head frozen. Inside Cutie on DAVIS 2017 val with the head and protocol of Table~\ref{tab:davis_s} (30 videos, 61 objects; dense 87.26, ours 87.07 at 0.462 FLOPs), \jf{} and FLOPs, with our method minus the row in brackets: G2TM at threshold 0.94, 0.96, 0.98: 78.26 at 0.414 ($+8.81$ \ci{+5.92}{+11.89}), 86.16 at 0.595 ($+0.91$ \ci{-0.04}{+1.72}) and 87.35 at 0.791 ($-0.28$ \ci{-0.94}{+0.20}); DTEM at r64, r96, r128: 87.38 at 0.721 ($-0.31$ \ci{-1.00}{+0.17}), 87.43 at 0.594 ($-0.37$ \ci{-1.11}{+0.16}) and 84.00 at 0.480 ($+3.07$ \ci{+0.95}{+6.38}). The two baselines reach the method's accuracy only at 28\% to 71\% more FLOPs, and at about its FLOPs they are below it.

\textbf{Baselines on the held-out long videos.} The same three heads and 140 videos as Table~\ref{tab:long} (reference implementation, no schedule; the method is 0.388 of dense FLOPs, plain ALGM 0.98 is 0.459), our method minus the comparator in \jf{} (95\% intervals, matching the sign convention above): G2TM at threshold 0.94 is $+15.53$ \ci{+12.62}{+18.65} at 0.319 FLOPs, and at 0.98 it is $+1.54$ \ci{-0.17}{+3.23} at 0.596 FLOPs (on the 42 objects in videos over 600 frames $+0.74$ \ci{-2.75}{+4.15}, at 0.571 against 0.366). ALGM given the method's FLOPs, by lowering its threshold, loses accuracy: our method minus ALGM 0.96 is $+6.14$ \ci{+4.02}{+8.20} at 0.383 FLOPs (ALGM 0.96 is $-3.23$ \ci{-4.81}{-1.71} against ALGM 0.98), and minus ALGM 0.95 is $+9.25$ \ci{+6.97}{+11.42} at 0.356 FLOPs. At or below the method's FLOPs no baseline we ran reaches its accuracy, and the one that comes within two points (G2TM 0.98) needs 54\% more.

\textbf{Gain by video length.} Table~\ref{tab:length} splits the held-out objects by the length of their video (three ViT-S heads). The method's advantage over ALGM 0.98 is clear on videos up to 600 frames ($+3.60$ up to 300 frames and $+2.95$ from 301 to 600) and not significant above 600 ($+0.73$ \ci{-3.37}{+5.53}); the permanent-memory schedule of Appendix~\ref{app:perm} gains more the longer the video, on every setting. The two patterns differ in where the point estimate peaks: without the schedule the gain over ALGM is largest up to 300 frames ($+3.60$), and with the same schedule in both it is largest from 301 to 600 frames ($+4.23$); above 600 frames it is not significant in either ($+0.73$ and $+1.42$).

\begin{table}[htbp]
\centering
\small
\caption{Held-out objects by video length (three ViT-S heads, per object; 95\% bootstrap intervals). Schedule: every 20th frame written to Cutie's permanent memory (Appendix~\ref{app:perm}).}
\label{tab:length}
\begin{tabular}{lccc}
\toprule
 & $\le$300 frames & 301--600 & $>$600 \\
\midrule
Videos / objects & 58 / 128 & 52 / 69 & 30 / 42 \\
Ours $-$ ALGM 0.98 & $+3.60$ \ci{+1.23}{+6.08} & $+2.95$ \ci{+0.03}{+5.91} & $+0.73$ \ci{-3.37}{+5.53} \\
Ours $-$ dense & $-0.66$ \ci{-2.23}{+1.11} & $-1.53$ \ci{-4.18}{+1.08} & $-0.29$ \ci{-3.80}{+3.64} \\
Schedule gain, dense Cutie & $+1.55$ \ci{+0.31}{+2.79} & $+2.86$ \ci{+1.35}{+4.68} & $+9.99$ \ci{+6.20}{+14.22} \\
Schedule gain, ALGM 0.98 & $+2.16$ \ci{+0.95}{+3.44} & $+2.31$ \ci{-0.16}{+4.77} & $+11.06$ \ci{+6.25}{+16.04} \\
Schedule gain, ours & $+2.04$ \ci{+0.41}{+3.71} & $+3.60$ \ci{+0.87}{+6.37} & $+11.75$ \ci{+6.45}{+16.62} \\
Ours $-$ ALGM, equal schedule & $+3.48$ \ci{+1.34}{+5.85} & $+4.23$ \ci{+1.31}{+7.49} & $+1.42$ \ci{-0.95}{+4.21} \\
Ours $-$ dense, equal schedule & $-0.17$ \ci{-1.34}{+1.15} & $-0.79$ \ci{-2.97}{+1.45} & $+1.48$ \ci{-1.65}{+4.37} \\
\bottomrule
\end{tabular}
\end{table}

\section{Settings}
ViT-S: high tier full resolution; middle tier under ALGM 0.98's rules; low tier budget at nominal rate 0.7 (capped at one half per block by pairing), floor 64 tokens; margin 1 patch; base map rounded up to 64-patch size classes; fallback plain ALGM 0.98. ALGM merges locally in layer 1 and globally in layer 5 (ViT-S, ViT-B) or 7 (ViT-L). ViT-B: ALGM 0.95, floor 128, margin 2. ViT-L: ALGM 0.95, floor 128, margin 1. Expectation and surprise settings are pixel-based and identical across backbones: block-matching radius 8 patches on $4\times4$-pooled pixels, camera search within 4 patches, variance floor 0.05 (with $10^{-5}$ added), the best of nine offsets in $\{-4,0,4\}^2$ pixels, threshold 0.06, frame-wide rule when more than half of the patches outside the base map fire, object masks at 1\% patch coverage and templates at 50\%, velocity shifts rounded to whole patches, lost below a tenth of the median of the last 10 tracked areas or 16 pixels.

\end{document}